%% file: t-its2023-nerf.tex
\documentclass[lettersize, journal]{IEEEtran} 
\usepackage{times}
\usepackage{amsmath}
\usepackage[pdftex]{graphicx}
\usepackage{subfigure}
\usepackage{subfiles} 
\usepackage{amsmath,amssymb,amsopn,amstext,amsfonts}
\usepackage{cancel}
\usepackage[space]{cite}
\usepackage{pdfsync}
\usepackage{balance}
\usepackage{color}
\usepackage{algorithm}
\usepackage{algorithmic}
\usepackage{url}
\usepackage{booktabs}
\usepackage{threeparttable}
\usepackage[colorlinks, citecolor=blue]{hyperref}
\usepackage{multirow}
\usepackage[outdir=./epstopdf/]{epstopdf}
\usepackage{graphicx}
\usepackage{array}
\usepackage{verbatim}
\usepackage{makecell}
\usepackage{balance}
\IEEEoverridecommandlockouts

\definecolor{brown}{rgb}{0, 0, 0}
\definecolor{lblue}{rgb}{0, 0, 0}

\newcommand{\revise}[1]{\textcolor{brown}{#1}}
\newcommand{\revv}[1]{\textcolor{brown}{#1}}

\graphicspath{{./figure/}}

\DeclareGraphicsExtensions{.pdf,.png,.jpg,.eps,.PNG}

\title{Crowd-Sourced NeRF: Collecting Data from Production Vehicles for \revv{3D Street View Reconstruction}}
\author{Tong Qin, Changze Li, Haoyang Ye, Shaowei Wan, Minzhen Li, Hongwei Liu,  and Ming Yang
	
	\thanks{This work is supported in part by the National Natural Science Foundation of China under Grants U22A20100 and 62173228
		(Corresponding author: Ming Yang; e-mail: MingYANG@sjtu.edu.cn).}
	\thanks{Tong Qin, Changze Li, and Ming Yang are with the Global Institute of Future Technology, Shanghai Jiao Tong University, Shanghai, China.}%
	\thanks{Haoyang Ye, Shaowei Wan, Minzhen Li, and Hongwei Liu are with IAS BU, Huawei Technologies, Shanghai, China.	}%
	}

\begin{document}

\maketitle

\subfile{sections/abstract_introduction}

\begin{figure*}
	\centering
	\includegraphics[width=0.95\textwidth]{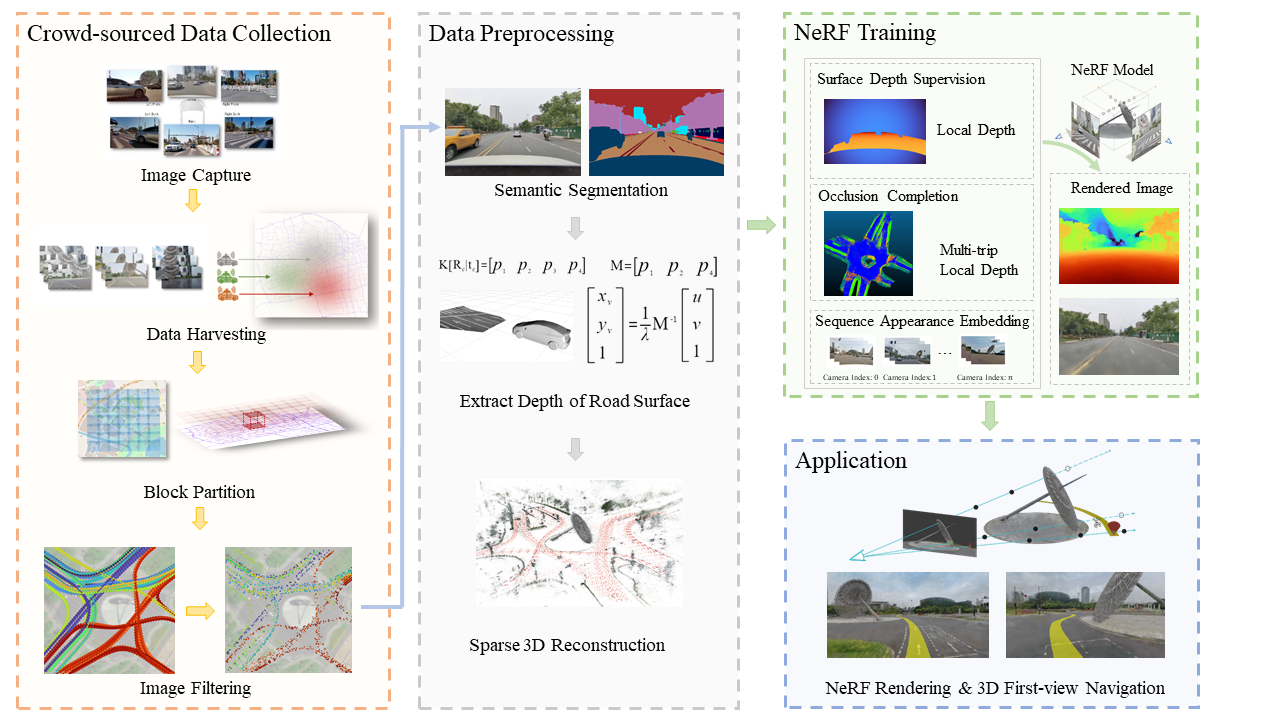}
	\caption{
		\revv{
		The structure of the proposed crowd-sourcing system. The strategy of crowd-sourced data collection is elaborated in Sec. \ref{sec:data_collection}, which collects massive data and filters them with a balanced spatial and temporal distribution.
			Then, the data pre-process model, Sec. \ref{sec:data_propocess}, segments images semantically, extracts the depth of the ground surface, and refines the camera pose by SfM. 
			The NeRF training procedure is illustrated in Sec. \ref{sec:trainning}, which trains the NeRF model with three improvements, which are sequence appearance embedding, surface depth supervision, and occlusion completion.
		}}
	\label{fig:framework}
\end{figure*}

\subfile{sections/literature_review}

\section{Framework Overview}
An overview of our method is shown in Fig. \ref{fig:framework}. 
The strategy of crowd-sourced data collection is elaborated in Sec. \ref{sec:data_collection}, which collects massive data and filters them with a balanced spatial and temporal distribution.
Then, the data pre-process model, Sec. \ref{sec:data_propocess}, segments images semantically, extracts the depth of the ground surface, and refines the camera pose by SfM. 
The NeRF training procedure is illustrated in Sec. \ref{sec:trainning}, which trains the NeRF model with three improvements, which are sequence appearance embedding, surface depth supervision, and occlusion completion.

\section{Crowd-Sourced Data Collection}
\label{sec:data_collection}

\subfile{sections/method_data_collection}

\subfile{sections/method_3d_reconstruction}

\section{NeRF Training}
\label{sec:trainning}
\subfile{sections/method_nerf_training}

\section{Experiments}
\label{sec:experiment}
In this section, we performed real-world experiments to evaluate the proposed system.
The data was captured in a crowd-sourcing way by multiple production vehicles.
We focused on the qualitative and quantitative results compared against other state-of-the-art algorithms.

\subfile{sections/experiment_data_collection}

\subfile{sections/experiment_reconstruction}

\subfile{sections/experiment_trips}


\subfile{sections/experiment_navigation}

\subfile{sections/method_rendering}

\subfile{sections/conclusion}


\bibliography{paper.bib}



\subfile{sections/biography}

\end{document}

%% file: sections/abstract_introduction.tex
\begin{abstract}
Recently, Neural Radiance Fields (NeRF) achieved impressive results in novel view synthesis.
Block-NeRF showed the capability of leveraging NeRF to build large city-scale models.
For large-scale modeling, a mass of image data is necessary.
Collecting images from specially designed data-collection vehicles can not support large-scale applications.
How to acquire massive high-quality data remains an opening problem.
Noting that the automotive industry has a huge amount of image data, crowd-sourcing is a convenient way for large-scale data collection.
In this paper, we present a crowd-sourced framework, which utilizes substantial data captured by production vehicles to reconstruct the scene with the NeRF model.
This approach solves the key problem of large-scale reconstruction, that is where the data comes from and how to use them.
Firstly, the crowd-sourced massive data is filtered to remove redundancy and keep a balanced distribution in terms of time and space.
Then a structure-from-motion module is performed to refine camera poses.
Finally, images, as well as poses, are used to train the NeRF model in a certain block.
\revv{
We highlight that we presents a comprehensive framework that integrates multiple modules, including data selection, sparse 3D reconstruction, sequence appearance embedding, depth supervision of ground surface, and occlusion completion. 
The complete system is capable of effectively processing and reconstructing high-quality 3D scenes from crowd-sourced data.
}
Extensive quantitative and qualitative experiments were conducted to validate the performance of our system.
\revv{
Moreover, we proposed an application, named first-view navigation, which leveraged the NeRF model to generate 3D street view and guide the driver with a synthesized video.
}

\end{abstract}

\begin{IEEEkeywords}
	Crowd-sourced system, intelligent vehicles, scene reconstruction, navigation, NeRF.
\end{IEEEkeywords}

\begin{figure}
	\centering
	\subfigure[The basic idea of crowd-sourced NeRF]{
		\includegraphics[width=.95\linewidth]{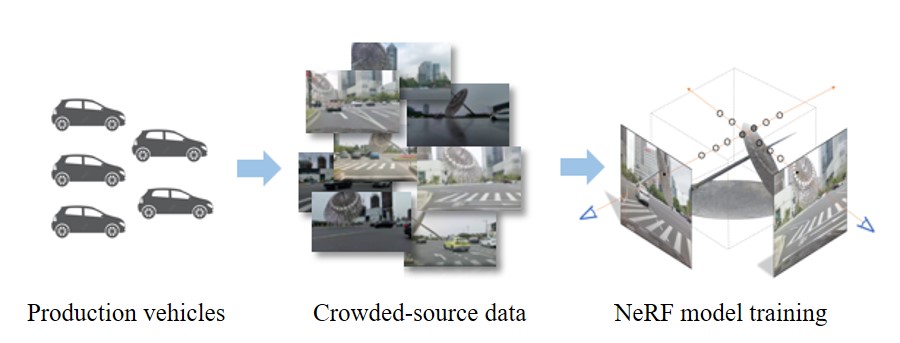} \label{fig:abs_a}
	}
	\subfigure[Synthetic view with navigation information.]{
		\includegraphics[width=.9\linewidth]{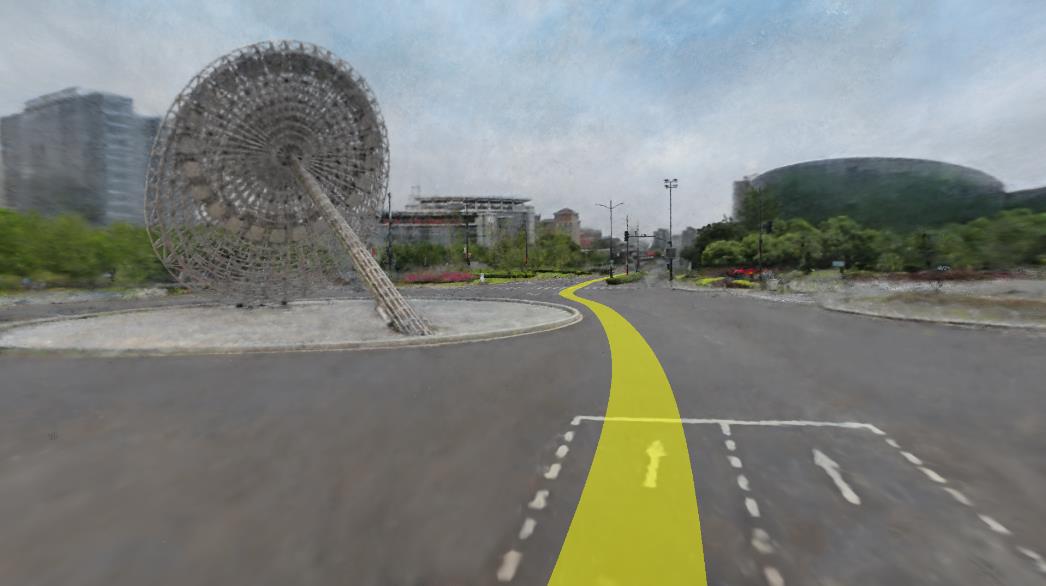} \label{fig:abs_b}
	}
	\caption{(a) \revv{shows the basic idea of crowd-sourced NeRF, which is collecting data from production vehicles to train the NeRF model for large-scale reconstruction. (b) shows an application, first-view navigation. A reference line (in yellow) is rendered with the realistic scene, which provides the driver with clearer experience.}
		\revise{The video can be found at: \url{https://youtu.be/oVUC634R1zw}.}}
	\label{fig:abs}
	\end {figure}
	
\section{Introduction}
\IEEEPARstart{S}{cene} reconstruction has been a long-standing topic over the last decades.
Classical Structure-from-Motion(SfM)-based method, such as COLMAP \cite{schoenberger2016sfm,schoenberger2016mvs}, reconstructed a 3D model from a set of 2D images taken from different angles. 
The basic idea was to find the 3D coordinates of points in the scene by triangulating corresponding points in multiple images. 
The resulting 3D points can then be used to estimate the camera poses and reconstruct the 3D geometry of the scene.
Besides the sparse 3D points, the dense surface could be stuck piece by piece. 
However, the canonical SfM approaches focused on 3D structures, ignoring photo-realistic textures.
Recently, a learning-based approach, named NeRF\cite{nerf}, represented the scene in the neural model implicitly.   
NeRF assumed that the scene can be represented as a continuous function that maps a 3D point to a color and opacity value. 
The model was trained using a set of images and corresponding camera poses, and the continuous function can be queried to render novel views of the scene from any desired viewpoint.
NeRF can achieve high-fidelity and realistic image synthesis, which can be widely applied to Virtual Reality(VR), Augmented Reality(AR), and simulation in autonomous driving industries.

While NeRF is an impressive technology, it requires a large amount of high-quality data to generate accurate and realistic 3D models.
Earlier works tended to focus on object-centric reconstruction within a small size, e.g. a table and a room.
Block-NeRF\cite{block_nerf} proposed a partitioning strategy that separated the city into multiple small blocks so that NeRF could be extended into city-scale modeling.
\revise{
Block-NeRF is indeed a groundbreaking approach in leveraging Neural Radiance Fields for large-scale city modeling. However, it inherently relies on data collected via specialized vehicles equipped with high-resolution cameras and precision localization hardware, which is not scalable for extensive urban environments due to high costs and logistical complexities. 
}
How to acquire massive and high-quality data to support NeRF training remains an open question for large-scale reconstruction.

Crowded sourcing is an efficient way to collect massive data in a short time and at a relatively low cost.
Nowadays, thousands of vehicles equipped with multiple cameras run everywhere and every day, which forms an ideal platform for crowd-sourced data collection.
In addition, the data stream of crowd-sourcing is stable and constant, which can support the daily updating of the model.
\revise{Therefore, Crowd-sourced data is cost-effective and has good real-time performance, which is crucial for large-scale urban 3D reconstruction and updates.}
However, how to use crowd-sourced data efficiently is a challenging issue.
On the one hand, with the continued growth of the crowd-sourced database, there is a lot of redundant data inside, which brings a huge burden to the storage and computation resources.
On the other hand, due to lacking a high-accurate positioning system, the pose of the image is inaccurate, which brings great difficulties to fusing them together.

To address the above-mentioned challenges, we propose \revise{CS-NeRF (Crowd-Sourced NeRF)}, which uses data captured by countless production vehicles to reconstruct the scene with the NeRF model.
Firstly, the massive crowd-sourced data is filtered by a spatial-temporal selector, which removes the redundancy and keeps a balanced spatial and temporal distribution.
The image is further segmented into multiple classes.
The dynamic scene is masked to improve accuracy in the SfM and NeRF training procedure.
\revise{We use inverse projection to estimate the depth of ground pixels, which serves as the depth for the supervision.}
Due to lacking a high-accurate localization device on vehicles, the pose of data is inaccurate and noisy.
We perform SfM to precisely localize each image.
The NeRF model is trained with depth supervision, occlusion completion, as well as sequence appearance embedding.
Last but not least, we apply the NeRF model in a real application, first-view navigation, which synthesis the scene with navigation information together for clear guidance.
\revise{The contributions of this paper are summarized as follows:}
\begin{itemize}
	\item \revise{We propose a comprehensive framework for large-scale NeRF reconstruction, that integrates multiple modules, effectively processing and reconstructing high-quality 3D scenes from crowd-sourced data.} 
	\item \revise{We propose three improvements within our CS-NeRF framework: sequence appearance embedding, ground surface supervision, and occlusion completion.}
	\item \revise{Real-world experiments with massive crowd-sourced data are conducted to validate the performance of the proposed system.}
	 
\end{itemize}


\revise{The crowd-sourced data used in the experiment is open-sourced for the benefit of the community.}
\revise{ \footnote{\url{https://drive.google.com/drive/folders/1AGxYzPJjceb4xEs3CeEwC5h4J_cVq_AX?usp=sharing} }
\footnote{
	 \url{https://pan.baidu.com/s/1fECS4ue8HKUclq3Zqi3UiQ?pwd=gzs5}}
 }


%% file: sections/literature_review.tex
\section{Related Work}
\subsection{Implicit Neural Model}
Traditional geometry-based 3D scene representations usually represent the scene as explicit models, such as point clouds \cite{agarwal2011building}, textured meshes \cite{romanoni2017mesh} or voxels \cite{seitz1997photorealistic}.
The sparsity and discretization attributes of these models make it difficult to render high-quality novel views.
NeRF \cite{nerf} and its subsequent works \cite{barron2021mip,barron2022mip} parameterized the positions and directions by multilayer perceptron (MLP) and composited the novel views using volumetric rendering.
This structure ensured the model in a smooth and continuous representation, which helped to generate more photo-realistic and higher quality results \cite{gao2022nerf}.

Combining image segmentation supervision, Semantic-NeRF \cite{zhi2021place} jointly encoded density and radiance along with scene semantics.
Semantic information in images can be further used to mask dynamic objects for static scene modeling, as mentioned in \cite{block_nerf,urf}.
Panoptic-NeRF \cite{kundu2022panoptic} further involved segmentations to decouple background and foreground as different networks to produce implicit panoptic models.
\revise{PanopticLifting \cite{siddiqui2023panoptic} lifts 2D panoptic labels to an implicit 3D volumetric representation, which is robust to label noise and has the potential to enhance semantic temporal stability.}

Accurate surface reconstruction was difficult for NeRF models to achieve since NeRF mainly focused on learning view synthesis results rather than scene geometry.
Implicit signed distance function (SDF) methods \cite{wang2021neus,yariv2021volume} are proposed to better disentangle the geometry and radiance, and ensure high-quality surface geometry,
In these approaches, volume density functions in the original NeRF were replaced by transformed learnable SDF functions.
However, for textureless scenes, depth supervision was necessary to ensure good surface geometry.
For example, DS-NeRF \cite{deng2022depth} and URF \cite{urf} supervised volume densities to fit the true depth distributions from sparse point clouds.
On the other hand, these models could provide denser coverage than raw point clouds, thanks to the image observations utilized during training.
\revise{
DS-NeRF uses sparse point clouds obtained from SfM for depth supervision, but this method is unable to triangulate a large number of reliable 3D points in texture-less structures such as the ground surface, which results in the geometry not being able to be improved effectively.
URF uses LiDAR for depth supervision, which may not be available in customer vehicles.
In contrast, Our proposed method uses inverse mapping to avoid the problems of the above methods.}

Instead of modeling synthetic or small indoor scenes, Block-NeRF \cite{block_nerf}, Urban Radiance Fields (URF) \cite{urf} targeted to real-world outdoor mapping applications.
\revv{These approaches took street-view images in individual blocks, and depths if available, as inputs to generate large-scale NeRF models.}
Similarly, Mega-NeRF \cite{turki2022mega} and BungeeNeRF \cite{xiangli2022bungeenerf} made use of aerial images or multi-scale satellite data.
\revise{
Although these methods can generate urban-level NeRF models, their data are collected using professional high-precision acquisition equipment, which reduces the complexity of the problem.
In contrast, the data collected in our method comes from crowd-sourced data, which is of low quality.
In order to get good reconstruction results, it is necessary to perform data cleaning and pose optimization, as well as add some constraints for crowd-sourced scenarios, such as scene appearance embedding.
}
By adding additional transient and appearance embeddings, NeRF-W \cite{Martin2021} addressed the appearance mismatches across different images.
To deal with unbounded environments, different scene parameterizations \cite{zhang2020nerf++,barron2022mip,turki2022mega} were also used in these methods.


\subsection{Camera Pose Optimization}
Accurate camera poses are essential for NeRF models to converge and obtain consistent color and occupancy.
Classical Structure-from-Motion (SfM) \cite{agarwal2011building,wu2013towards,schoenberger2016sfm} was an effective offline way to guarantee the accuracy of camera poses, as well as the sparse scene structure.
As proposed in \cite{nerf--,lin2021barf,block_nerf,iMAP,NICE-SLAM}, jointly optimizing camera poses and NeRF models during training can further improve the model consistency, meanwhile reducing the requirements for very accurate camera poses.
\revise{NoPe-NeRF \cite{bian2023nope} has utilized monocular depth estimation to generate dense point clouds and then used the constraints of the point clouds to optimize both the neural radiance field and the camera pose.
However, due to the accuracy of monocular depth estimation and the problem of local optima, it is difficult to achieve good results in large-scale autonomous driving scenarios.}

\revise{In our method, pose optimization is considered to compensate for calibration errors and noisy camera poses from customer vehicles.}

\subsection{Crowd-sourced Mapping}
Data collection by laboratory sensor settings, such as the one used in Block-NeRF \cite{block_nerf}, were usually not affordable for large-scale applications.
\revv{For instance, high-resolution camera with 2K resolution and high-accurate positioning system with centimeter-level accuracy.}
There exist several works related to crowd-sourced mapping for geometric models.
Crowd trajectory data was utilized in \cite{ruhhammer2016automated} for automated intersection mapping.
In \cite{kim2018crowd, qin2021light}, a crowd-sourced process was used for the new HD map feature layer generation.
For NeRF applications, NeRF-W \cite{Martin2021} was one of the earliest works related to crowd-sourcing.
It took internet photo collections as inputs and succeeded to produce a static model without transient elements.
To produce driving-view models, our approach intended to use data from production vehicles.
We will introduce the crowd-sourced NeRF system dealing with low-fidelity and noisy data in the following sections.

%% file: sections/method_data_collection.tex
\subsection{Data Collection Platform}
The data is collected in a crowd-sourcing way by multiple mass-produced vehicles.
Only \revise{an integrated positioning system} and cameras with intrinsic and extrinsic calibrations are required.
A large number of production vehicles, equipped with ADAS (Advanced Driver-Assistance Systems) or ADS (Automated Driving Systems), meet the hardware requirements and can be served as the data collection source.
Images, along with meter-level global poses, are uploaded to the data collection platform.

\subsection{Data Selection}
Since substantial data aggregates continuously, the data selection pipeline is designed to efficiently select images and reduce redundancy.
The data selection pipeline ensures the spatial and temporal distribution of the data by block partitioning and image filtering.

\subsubsection{Block Partition}
Following Block-NeRF\cite{block_nerf}, the scene is divided into small blocks.
The crowd-sourced data is assigned to blocks according to the pose.
The adjacent blocks are overlapped by 20\% to ensure consistency.
Subsequently, a separate neural network can be trained for each block, enabling parallel processing and reducing overall training time.

\subsubsection{Image Filtering}
\revise{
    In each block, images are filtered automatically according to following principles:
\begin{itemize}
	\item Small proportion of moving objects: We identify moving objects through semantic segmentation. 
	Images with a proportion of moving objects greater than 40 percent are \revise{filtered out}.
	\item Stable pose estimation: Theoretically, the prior pose should not deviate too far after optimization. 
	Therefore, if the pose of a specific image changes a lot after the pose optimization, the image is considered untrustworthy and is filtered out.
	\item Data diversity: Images captured with similar times, positions, and viewpoints can be clustered by utilizing optimized poses and timestamps. Only few images are retained in each clustering category. 
\end{itemize}
}

\section{Data Preprocessing}
\label{sec:data_propocess}

\begin{figure}
	\centering
	\subfigure[The example of semantic segmentation result.]{
		\includegraphics[width=.8\linewidth]{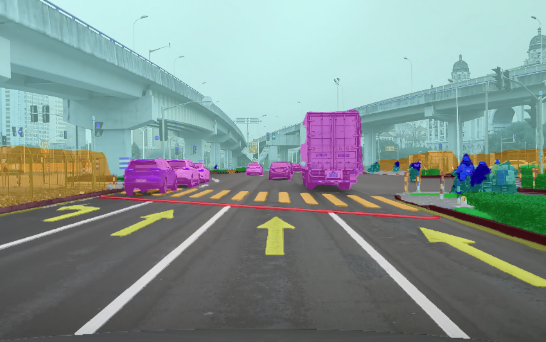} \label{fig:segmentation}
	}
	\subfigure[The illustration of ground surface projection.]{
		\includegraphics[width=.8\linewidth]{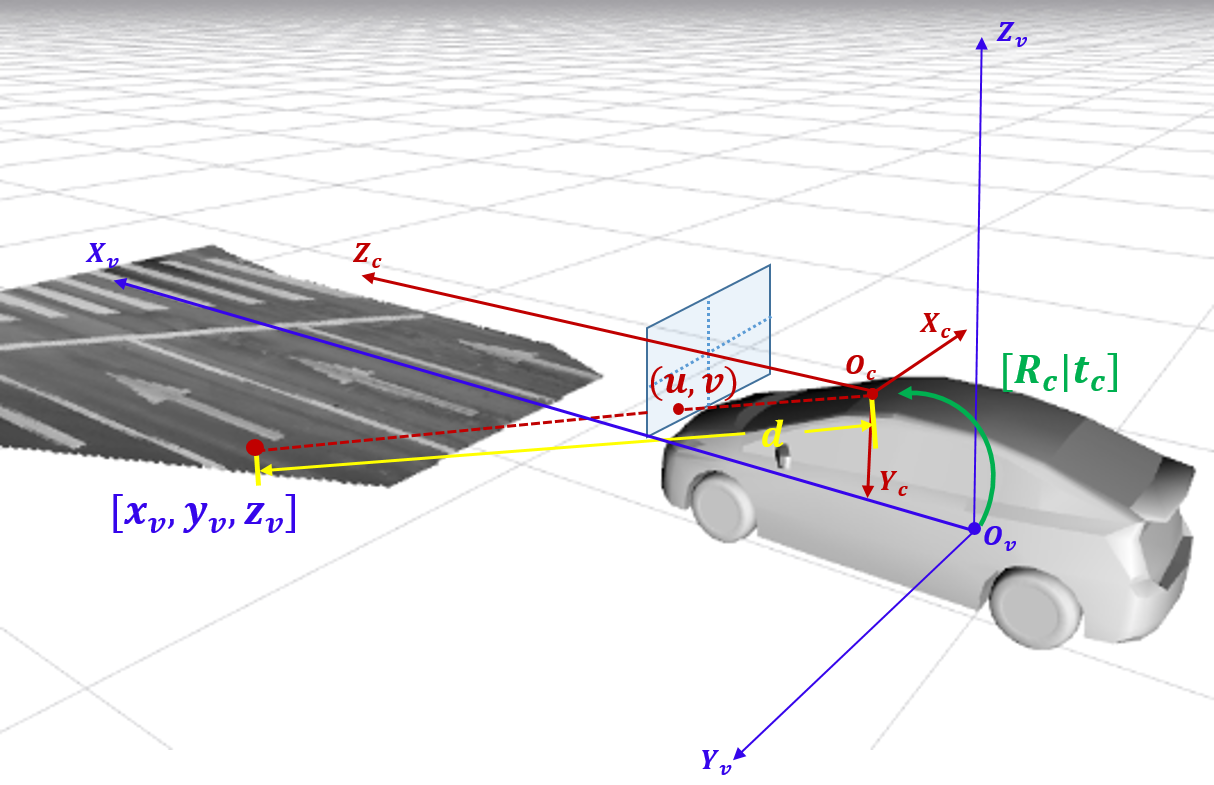} 
		\label{fig:imp_projection}
 	}
	\caption{In (a), the image is segmented into multiple semantic groups, such as lane, crosswalk, vehicle, tree, road, stop lines, etc. (b) is the diagram of the inverse projection process. The pixel is inversely projected to the ground ($z_v = 0$), so that the depth $d$ of the ray can be obtained.} 
	\label{fig:segmentation_projection}
\end{figure}

\subsection{Semantic Segmentation}
Semantic segmentation is a mature technology, which has been widely used in visual applications.
Typical CNN-based methods include FCN\cite{long2015fully}, U-Net \cite{ronneberger2015u}, SegNet\cite{badrinarayanan2015segnet}, BiSeNet V2\cite{yu2021bisenet}, etc.
In this project, we adopted BiSeNet V2\cite{yu2021bisenet}, which had a good trade-off between speed and accuracy.
In BiseNet V2, the network was separated into two branches.
The detail branch captured low-level details and generated high-resolution features, while the semantic branch with deep layers, obtained high-level semantic context.

Semantic segmentation was performed on each image.
As shown in Fig. \ref{fig:segmentation}, the image was segmented into multiple semantic groups, such as lane, crosswalk, vehicle, tree, road, stop lines, etc.
There are two usages for semantic segmentation results:
\begin{itemize}
	\item Mask moving object. The moving objects, such as cars and pedestrians, severely impact the accuracy of the following 3D reconstruction and NeRF model training.
	\item Road surface extraction. \revv{Through segmentation, the road surface can be detected on the image plane, which will be used for surface depth generation in the next section. }
\end{itemize}

\subsection{\revv{Depth of Ground Surface}}
\label{sec:depth_prior}
Since the data is captured on the open street, we assume that the road surface is a roughly flat plane. 
\revv{Therefore, the road surface can provide the depth for supervision in the NeRF training process.}
\revv{The depth of the road surface can be inferred by inverse projection.}
A diagram of the inverse projection process is shown in Fig. \ref{fig:imp_projection}.
The pixel is inversely projected to the ground ($z_v = 0$), so that the depth $d$ of the ray can be obtained.

Specifically, the formula for projecting a 3D point under the vehicle coordinates  $[x_v$, $y_v$, $z_v]$ onto the 2D image plane $[u$, $v]$ is:
\begin{equation}
\begin{bmatrix} u \\ v \\ 1 \end{bmatrix} = \lambda \mathbf{K} 
\begin{bmatrix} \mathbf{R}_c | \mathbf{t}_c \end{bmatrix} 
\begin{bmatrix} x_v \\ y_v \\ z_v \\ 1 \end{bmatrix},
\end{equation}
where $\mathbf{R}_c$, $\mathbf{t}_c$ is the extrinsic calibration of the camera with respect to the vehicle’s center,
$\mathbf{R}_c \in SO(3)$.
$\mathbf{K}$ is the intrinsic parameter of the camera, $
\mathbf{K} = 
\begin{bmatrix}
f_x, 0, c_x\\
0, f_y, c_y \\
0, 0, 1
\end{bmatrix}	$,
where $f_x, f_y$ are focal lengths, and $c_x, c_y$ represent the camera principal point.
This equation projects the 3D point to 2D pixel $[u$, $v]$ on the image plane.
$\lambda$ is a scaling factor.
We define the origin of the vehicle coordinates as located on the ground, so $z_v$ is 0 for points on the ground.
$x_v$ and $y_v$ can be solved given $\mathbf{K}$, $\mathbf{R}_c$, $\mathbf{t}_c$,  $[u$, $v]$ and $z_v = 0$.
Therefore, we can get the depth $d$ of the ray, which is the distance between the 3D surface point $[x_v, y_v, 0]$ and the camera's center.
The depth will be used for depth supervision in Sec. \ref{subsec:depth_supervision}.


%% file: sections/method_3d_reconstruction.tex

\subsection{Sparse 3D Reconstruction}
The crowd-sourced data is in meter-level positioning accuracy, which is insufficient for NeRF training.
To address this issue, we perform Structure-from-motion (SfM) to further improve localization accuracy.
\revv{We adopt COLMAP \cite{schoenberger2016sfm}, which refines the geometric structure of the scene by feature extraction, feature matching, and bundle adjustment (optimizing camera poses and feature positions).}
Some improvements are made to improve accuracy in the feature matching step:
\begin{itemize}
	\item Since invalid features from dynamic objects (car, pedestrian, bicycle, etc.) affect the accuracy of the 3D scene's reconstruction, we remove features on dynamic objects based on the segmentation.
	\item We use semantic labels to further filter out mismatched pairs. 
	Only pairs with the same semantic label are kept.
	For example, features belonging to the road, or the building are treated as inliers.
\end{itemize} 
In addition, some improvements are made to improve \revise{the effectiveness of the SfM pose optimization process}:
\begin{itemize}
	\item Since we have the meter-level position for each image, the prior pose can guide feature matching instead of exhaustively searching.
	We perform feature matching only with candidate images within the neighborhood.
	\item \revise{In the bundle adjustment, the prior poses are adopted to guarantee the correctness of optimization and accelerate convergence.}
\end{itemize} 
Therefore, we have a centimeter-level position for crowd-sourced images, which can be used for further NeRF training.

%% file: sections/method_nerf_training.tex
\subsection {NeRF Preliminaries}
Neural Radiance Field (NeRF) represents the scene as a \revise{continuous} function, whose input is a 5D vector, including a 3D position $\mathbf{x} = (x, y, z)$ and a view direction $\mathbf{d} = (\theta, \phi)$, and whose output is an emitted color $\mathbf{c} = (r, g, b)$ and volume density $\mathbf{\sigma}$.
\revise{In order to improve the high-frequency detail of the rendered images, NeRF incorporates positional encoding:
    \begin{equation}
        \gamma \left( \mathbf{x} \right) = {\left[ {\sin \left( \mathbf{x} \right),\cos \left( \mathbf{x} \right), \ldots ,\sin \left( {{2^{L - 1}}\mathbf{x}} \right),\cos \left( {{2^{L - 1}}\mathbf{x}} \right)} \right]^T}
    \end{equation}
    where $L$ is a hyperparameter.
The function can be described as an MLP network $\mathbf{F_\Theta}:(\gamma \left(\mathbf{x}\right), \mathbf{d}) \rightarrow (\mathbf{c}, \mathbf{\sigma})$\cite{nerf}.
It is worth noting that $\mathbf{x}$ in the NeRF network is inputted at the first layer, while the $\mathbf{d}$ is injected into the network closer to the end of the MLP to alleviate the "shape-radiance ambiguity" \cite{zhang2020nerf++}.}
Colors can be obtained by volume rendering of any ray passing through the scene, the expected color $C(\mathbf{r})$ of the ray  $\mathbf{r}(t) = \mathbf{o} + t \mathbf{d}$ within near and far bounds $[t_n,t_f ]$ is:
\begin{equation}
	\label{eq:color_render}
	C(\mathbf{r}) = \int_{t_n}^{t_f} w(t) \mathbf{c}(\mathbf{r}(t), \mathbf{d})dt,
\end{equation}
where $w(t) = \exp \left( -\int_{t_n}^{t}\sigma(\mathbf{r}(s)) ds \right) \cdot \sigma(\mathbf{r}(t))$.
The training process is supervised by an $L2$ photometric loss:
\begin{equation}
	\label{eq:rgb_loss}
	{L}_{rgb} =  \Vert C_i(\mathbf{r})-C_i^{gt}(\mathbf{r})\Vert_2^2 ,
\end{equation}
where $C_i^{gt}(\mathbf{r})$ is the ground truth color of a pixel $i$ in the image.

We build on the method, Nerfacto\cite{nerfstudio}, which is an open-sourced toolbox for NeRF development, integrating a lot of new features.
Nerfacto integrated pose refinement from NeRF-\,- \cite{nerf--}, proposal network sampler and scene contraction from Mip-NeRF 360\cite{barron2022mip}, appearance embedding from NeRF-W\cite{Martin2021}, hash coding and fused MLP from Instant-NGP \cite{instant-ngp}.
In the following sections, we only elaborate on our modifications and improvements based on the Nerfacto.

\subsection{Sequential Appearance Embedding}
Crowd-sourced data is captured by the different vehicles at different times.
The inconsistency of image styles may lead to poor reconstruction performance.
Traditional algorithms, such as NeRF-W\cite{Martin2021}, assigned every image with an appearance embedding vector.
Although the appearance embedding solved the ambiguity of reconstruction, the high degree of freedom causes the model to learn some randomness in a single image, such as shadows.
Instead of assigning appearance embedding vectors for each image, we assign appearance embedding vectors for each sequence.
\revise{In other words, the same sequence (images captured by the same camera on the same trip) shares the same embedding vector instead of using one appearance embedding vector per image.}
Therefore, we reduce the dimension of appearance embedding and reduce the randomness.
The model learns the common style for a series of images, instead of learning a special pattern for a certain image. 

\begin{figure}
	\centering
	\includegraphics[width=0.4\textwidth]{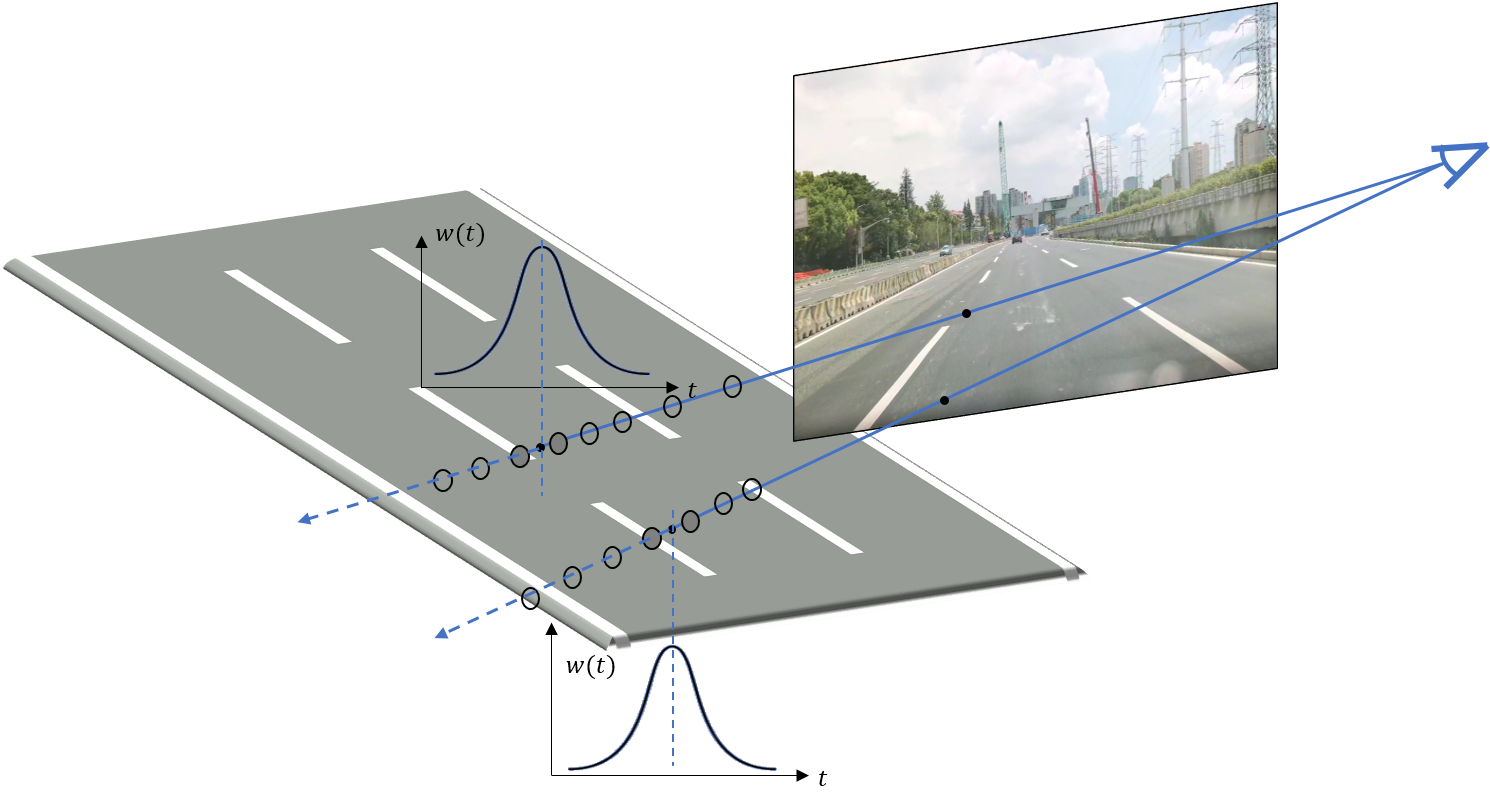}
	\caption{The illustration of depth supervision. The density distribution of sample points along a ray is supervised by the Dirac function.}
	\label{fig:depth_loss}
\end{figure}

\subsection{Depth Supervision of Ground Surface}
\label{subsec:depth_supervision}
In the canonical NeRF training procedure, the color of pixels is supervised while the geometry of the scene is neglected.
Due to the lack of geometric supervision, there are often many floating artifacts in the novel view \revise{\cite{urf, deng2022depth}}.
Depth supervision contributes to improving NeRF's geometric quality when the number of input views is limited.
\revv{As illustrated in Sec. \ref{sec:depth_prior}, the depth of the ray, which hits on the ground surface, is generated.
This depth is used for supervision to improve the geometric quality.}
We follow the depth supervision method proposed in Urban Radiance Fields (URF) \cite{urf}, by supervising the density distribution of a ray looking like the Dirac function, as shown in Fig. \ref{fig:depth_loss}. 
The depth loss function is defined as:
\begin{equation}
	\label{eq:sight_depth}
	{L}_{d} = \int_{t_n}^{t_f} (w(t) - \delta(z))^2\,dt ,
\end{equation}
where $\delta(z)$ is the Dirac function.
The final loss is the sum of color loss for all rays and depth loss for rays $\mathbf{s}$, which hits the road surface:
\begin{equation}
	\label{eq:fianl_loss}
	{L} = \sum_i {L}_{rgb}+  \sum_{s\in \mathbf{S}} {L}_{d}.
\end{equation}
After adding the depth loss, the rendered depth is smoother than the one without depth supervision, as shown in Fig. \ref{fig:render_quality}.

\subsection{Occlusion Completion}
Although semantic segmentation is used to mask dynamic objects (e.g. cats, pedestrians, bicyclists), masked areas may cause black holes due to a lack of efficient supervision.
As shown in Fig. \ref{fig:occlusion_complete}, black shadows appear in the area shaded by vehicles in the rendered view.
To this end, \revv{we fill depth of the ground in the mask area with the nearby ground.}
Therefore, the masked area is supervised effectively, and the quality of ground reconstruction is improved, as shown in Fig. \ref{fig:occlusion_complete}.

 \bigskip

Overall,  the comparison of proposed method against typical NeRF methods, such as Mip-NeRF\cite{barron2021mip}, 	Instant-NGP\cite{instant-ngp}, 	Nerfacto\cite{nerfstudio} ,	Block-NeRF\cite{block_nerf}, is shown in Table \ref{tab:methods_comparison}.

\begin{table*}[t]
	\setlength\tabcolsep{4.5pt}
	\centering
	\caption{{\revv{Comparison for different NeRF methods}}
		\label{tab:methods_comparison}}
	\setlength{\tabcolsep}{1mm}
	\begin{tabular}{l|c|c|c|c|c|c}
		\toprule
		\quad & Scene Contraction     & Hash Encoding  & Appearance Embedding  & Block Partition &  Depth Supervision & Occlusion Completion\\
		\midrule
		Mip-NeRF\cite{barron2021mip} &\checkmark & & &  &   \\
		Instant-NGP\cite{instant-ngp} &  &\checkmark   &  &  & \\
		Nerfacto\cite{nerfstudio}  & \checkmark & \checkmark & \checkmark & &  &  \\
		Block-NeRF\cite{block_nerf} & \checkmark &  & \checkmark & \checkmark & &  \\
		CS-NeRF(ours) & \checkmark & \checkmark & \checkmark & \checkmark & \checkmark & \checkmark \\
		\bottomrule
	\end{tabular}
	\begin{tablenotes}
		\footnotesize
		\item *Our method was built on the top of Nerfacto\cite{nerfstudio}.
	\end{tablenotes}
\end{table*}

%% file: sections/experiment_data_collection.tex

\subsection{Experimental Data Collection}
\subsubsection{Hardware Configuration}

\begin{figure}[t]
	\centering
	\subfigure[\revv{Commercial vehicle: ArcFox Alpha S}]{
		\includegraphics[width=.8\linewidth]{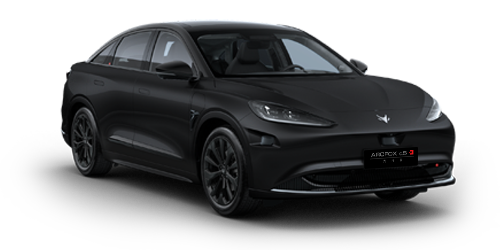} \label{}
	}
	\subfigure[Original sensor configuration]{
		\includegraphics[width=.6\linewidth]{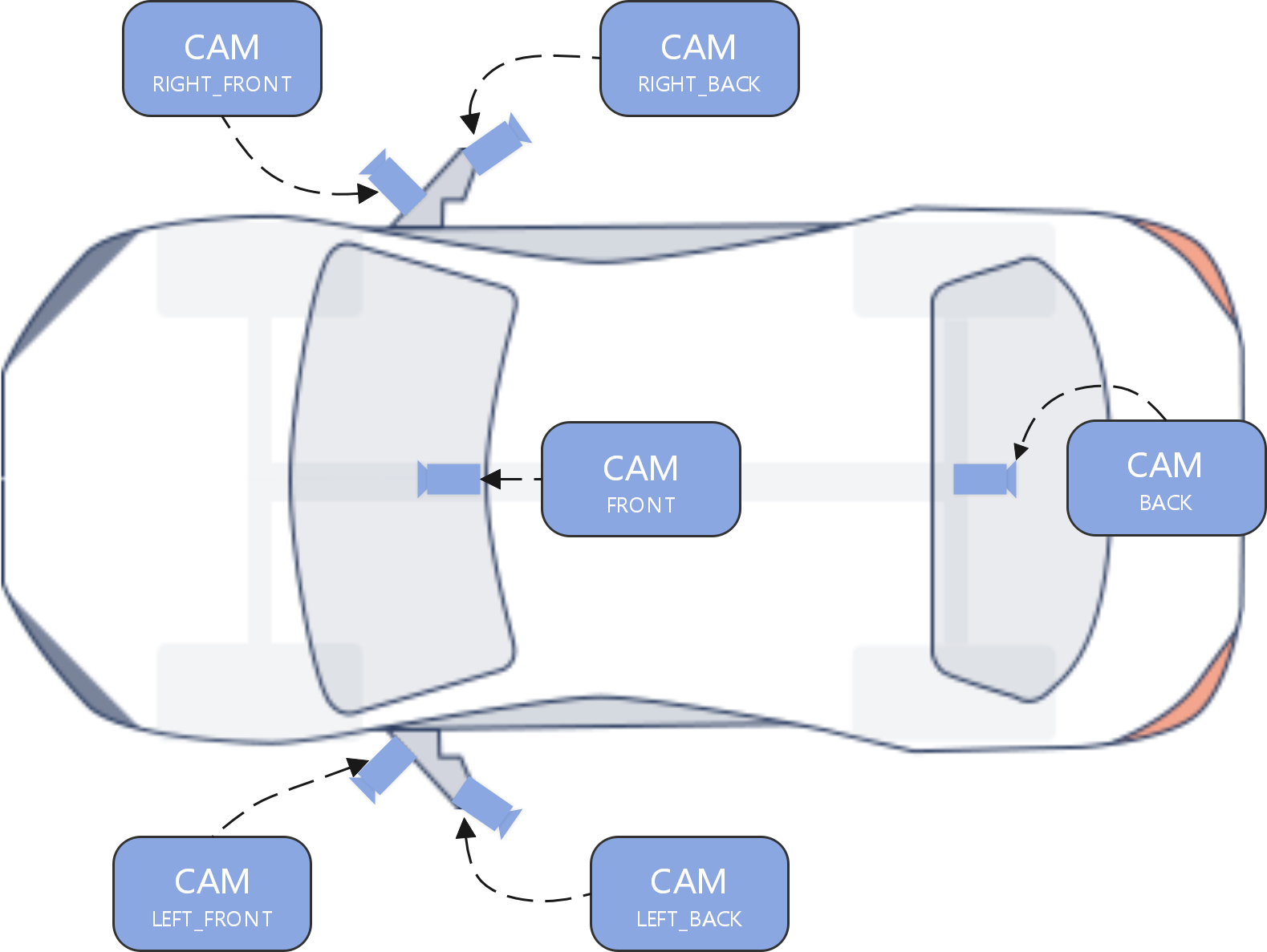} \label{}
	}
	\caption{\revv{(a) shows the vehicle we used for crowd-sourced data collection.
		(b) showns the sensor setup we used for experiments. (The vehicle contains more sensors than we used.)}}
	\label{fig:sensor_setup}
	\end {figure}

The dataset was collected on public urban roads by production vehicles, which were not specifically designed for data collection.
By cooperating with the automobile company, \textit{ArcFox Alpha S}, were used for data collection.
\revv{Our collaboration with the company operating these vehicles allowed us access to raw data directly from the fleet, encompassing a wide variety of users, locations, and environmental conditions.}
The vehicle was equipped with multiple cameras.
Among these cameras, 6 cameras were used in our project:
one located at the front window shield, one located at the back, and four located at the right and left side respectively, as shown in Fig. \ref{fig:sensor_setup}.
\revv{Specifically, cameras operate at 20Hz frequency with 640x480 resolution and JPEG compression. }
An example of captured images from one vehicle was shown in Fig. \ref{fig:platform}.
All cameras were factory-calibrated.
An integrated positioning system was incorporated with GPS, IMU, and wheel speedometer, which achieved meter-level localization accuracy.
\revv{Specifically, in open area, where GPS signals are well received, the localization accuracy can reach 1 meter, while in complex region, where GPS signals is occluded and reflected, the localization error is over 10 meters.}
One vehicle was specially equipped with Lidar to provide accurate depth only for the evaluation purpose. 
The images were captured on-board and uploaded to the cloud platform.
The proposed system run on the cloud service offline.

\revise{
}

\revise{
}
\subsubsection{Century Avenue Dataset}
\begin{figure}[t]
	\centering
	\subfigure[Images captured by the vehicle from six cameras.]{
		\includegraphics[width=.8\linewidth]{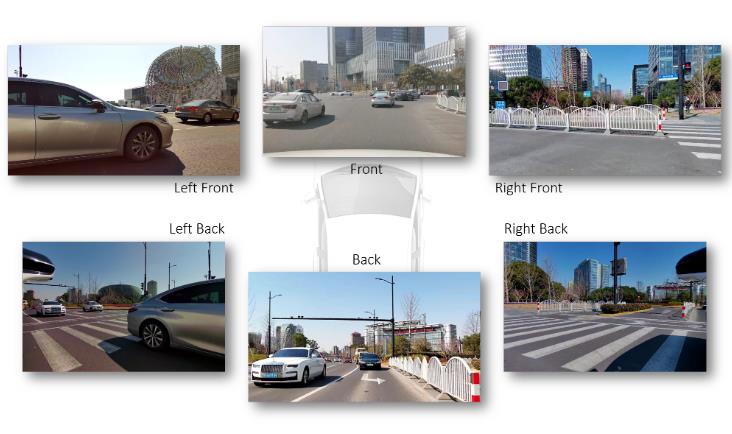} \label{fig:platform}
	}
	\subfigure[Sample images from crowd-sourced dataset.]{
		\includegraphics[width=.8\linewidth]{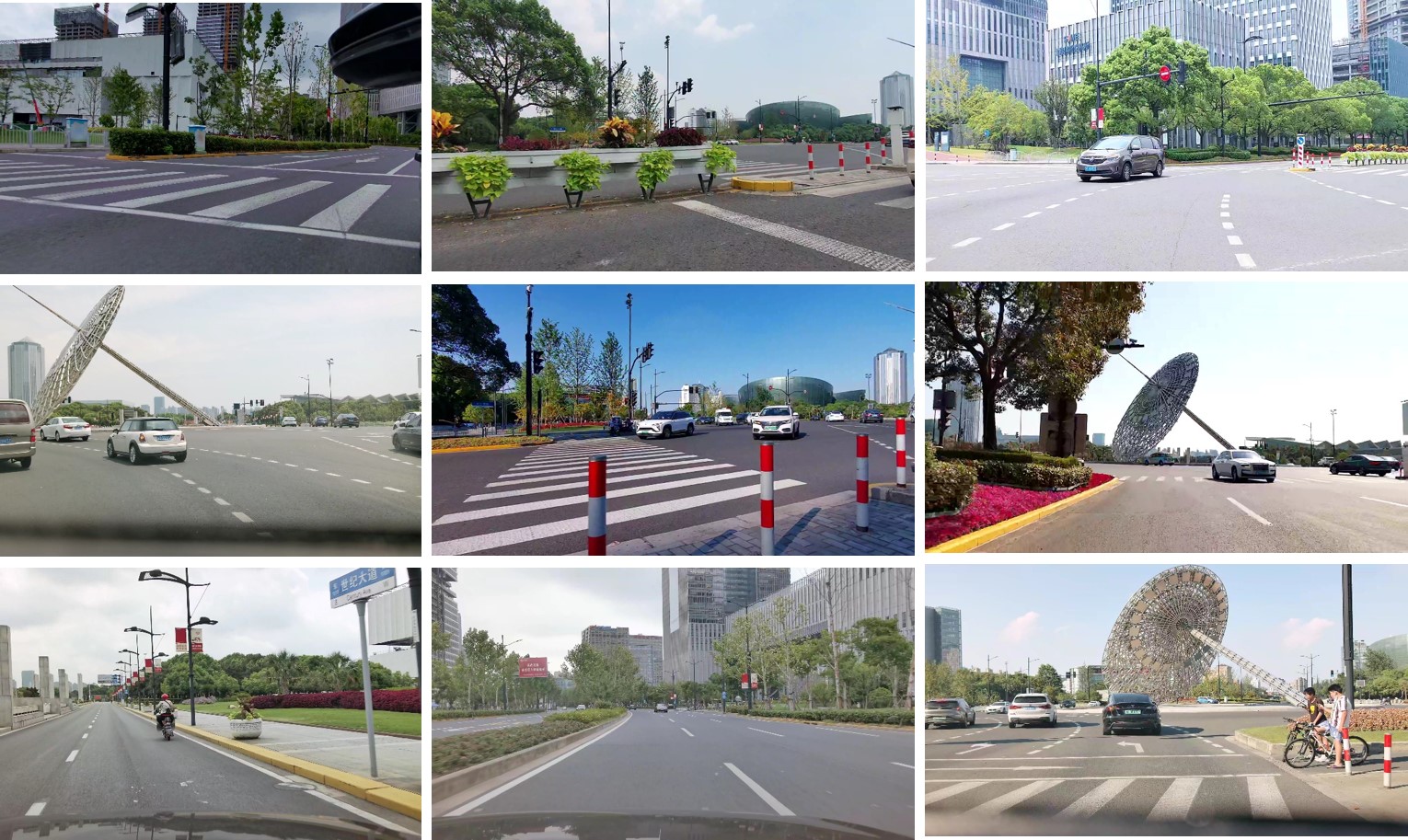} 
		\label{fig:sample_images}
	}
	\caption{ (a) shows images captured by the data collection platform. (b) shows the diversity of our crowd-sourced dataset.}
	\label{fig:Centry Avenue Dataset}
\end{figure}

\begin{figure}[t]
	\centering
	\subfigure[Data before filtering.]{
		\includegraphics[width=.45\linewidth]{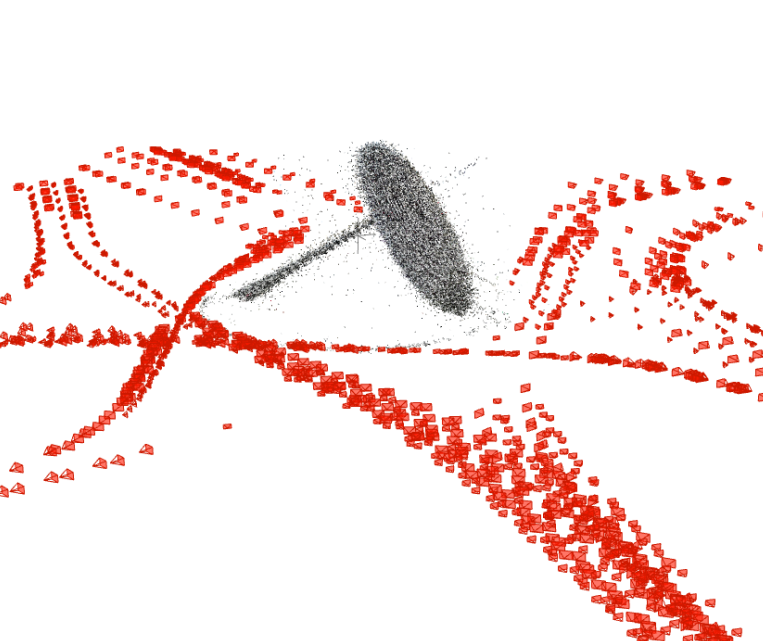} \label{}
	}
	\subfigure[Data after filtering.]{
		\includegraphics[width=.45\linewidth]{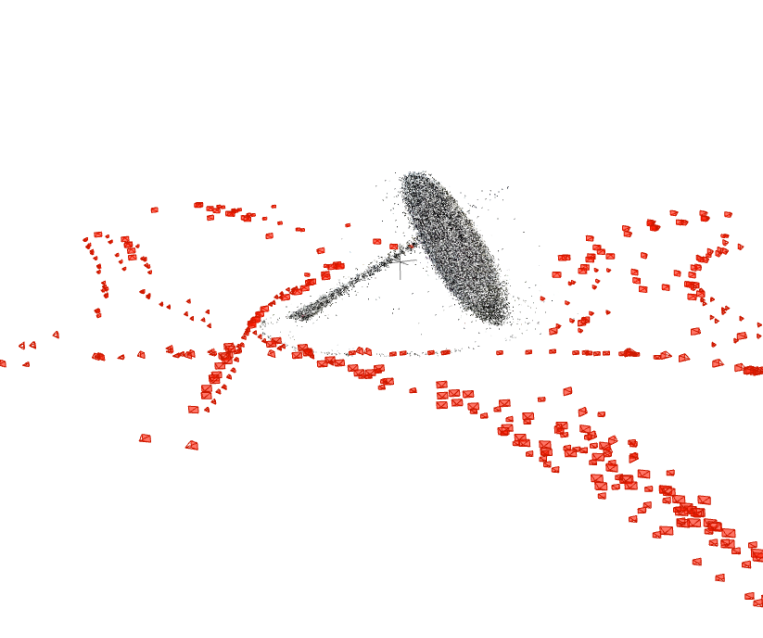} \label{}
	}
	\caption{The data selection procedure efficiently reduced the redundancy of the crowd-sourced data, while keeping a balanced spatial distribution.}
	\label{fig:data_filter}
	\end {figure}

In the following experiment, we focused on one block, the Century Avenue Intersection in Shanghai, China.
Driving in this area was challenging for human drivers due to its complex road structure. 
The block was covered  $150m \times 150m$. 
The crowd-sourced data contained $58$ trips passing through the area from February 2022 to August 2022, including winter, spring, and summer seasons. 
The local time spans from 9:00 am to 7:00 pm. 
The total data collection time was $1.06$ hours, with $29,732$ images.
Sample images from the crowd-sourced dataset were shown in Fig. \ref{fig:sample_images}, which illustrated the diversity of the data.
As described before, data needed to be filtered to reduce redundancy and ensure the efficiency of further processing.
Ultimately, $2,896$ images with balanced spatial and temporal distribution were selected.
An illustration of the dataset before and after filtering was shown in Fig. \ref{fig:data_filter}.

%% file: sections/experiment_reconstruction.tex
\subsection{Reconstruction Comparison} \label{sec:reconstruction_comparison}

\begin{table}[t]
	\setlength\tabcolsep{4.5pt}
	\centering
	\caption{{Metric comparison for different NeRF methods}
		\label{tab:method_comparison_experiment}}
	\setlength{\tabcolsep}{1mm}
	\begin{tabular}{lccccc}
		\toprule
		\quad & PSNR $\uparrow$ & SSIM $\uparrow$ & LPIPS $\downarrow$ & \begin{tabular}[c]{@{}c@{}}RMSE{[}m{]}\\ $@1\sigma$ \end{tabular} $\downarrow$ & \begin{tabular}[c]{@{}c@{}}RMSE{[}m{]}\\ $@2\sigma$ \end{tabular} $\downarrow$ \\
		\midrule
		Mip-NeRF\cite{barron2021mip} & 17.33 & 0.543 & 0.629 &  \underline{10.70} &  \underline{14.28} \\
		Instant-NGP\cite{instant-ngp} & 16.25  &  0.599 & 0.496 & 15.79 & 25.19 \\
		Nerfacto\cite{nerfstudio} &  \underline{21.33} &  \underline{0.748} & 0.193 & 14.13 & 23.84 \\
		CS-NeRF(ours) & \textbf{22.54} & \textbf{0.754} & \textbf{0.185} & \textbf{6.11} & \textbf{11.34}  \\
		\bottomrule
	\end{tabular}
	\begin{tablenotes}
		\footnotesize
		\item[1] *Comparisons of different NeRF method in the century avenue dataset. 
		\item[2] *Our method was built on the top of Nerfacto\cite{nerfstudio}.
	\end{tablenotes}
\end{table}

\begin{figure*}[h]
	\centering
	\includegraphics[width=1\textwidth]{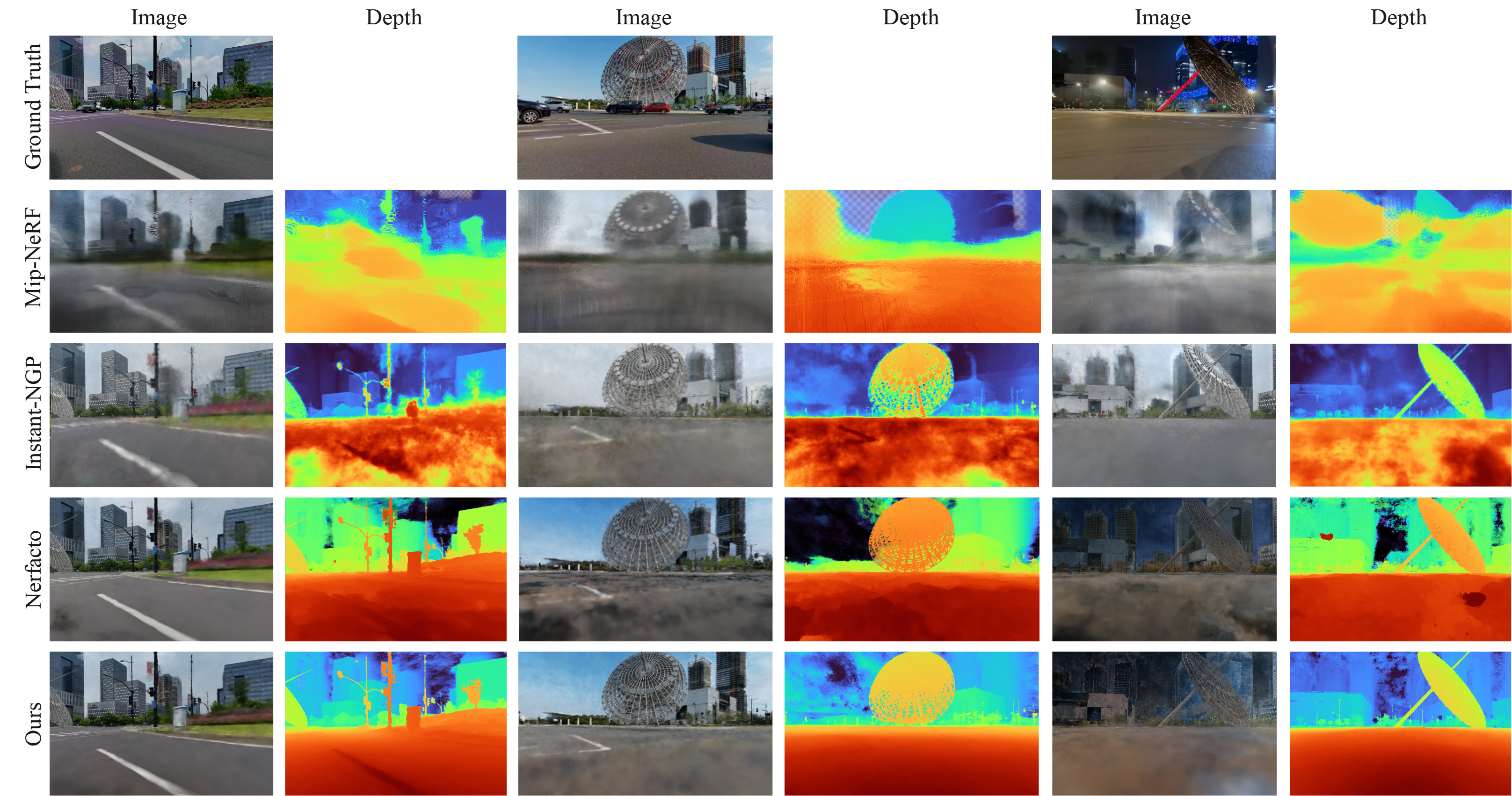}
	\caption{
		\revise{
		Figures shows qualitative results from experiments on the crowd-sourced dataset. Compared against Mip-NeRF\cite{barron2021mip}, Instant-NGP\cite{instant-ngp}, Nerfactor\cite{nerfstudio}, our method added three improvements, ground surface supervision, occlusion completion, and sequence appearance embedding. It can be seen that the depth of our approach is more accurate, the ground is smoother.
	}}
	\label{fig:render_quality}
\end{figure*}

We reported PSNR, SSIM\cite{SSIM}, and LPIPS\cite{LPIPS} metrics for novel view rendering.
\revise{
These three metrics evaluate the rendering quality of NeRF from different perspectives, which include image reconstruction errors and perceptual differences.
The calculation method can be found in \cite{SSIM} and \cite{LPIPS}.}
Not only evaluating appearance, but we also used RMSE (Root Mean Square Error) of depth to evaluate the geometrical reconstruction quality.
The ground truth of depth came from Lidar point clouds.
We compared our proposed method against Mip-NeRF\cite{barron2021mip}, Instant-NGP\cite{instant-ngp},and Nerfacto\cite{nerfstudio}.
Nerfacto\cite{nerfstudio} is an open-source implementation, which integrated multiple advanced features, such as pose refinement from NeRF-\,-\cite{nerf--}, proposal network sampler and scene contraction from Mip-NeRF 360\cite{barron2022mip}, appearance embedding from NeRF-W\cite{Martin2021}, hash coding and fused MLP from Instant-NGP \cite{instant-ngp}.
Our system was built on the top of Nerfacto.
Since Nerfacto optimized the appearance embedding vector for each image, there was no appearance embedding vector for test images.
For a fair comparison, we used the average of embedding vectors from the same camera as the embedding vector.
\revise{For each block, our training was performed on a single GPU (Nvidia Quadro P5000) with a training duration of approximately two hours.
Multiple blocks were processed in parallel.}
\revise{
The quantitative metric results were shown in Table. \ref{tab:method_comparison_experiment}.
It could be seen that our method outperforms others in terms of appearance and depth estimation.
Straightforwardly, figures of qualitative comparisons were shown in Fig. \ref{fig:render_quality}.
More detailed examples can be found in Fig. \ref{fig:surface} and Fig. \ref{fig:occlusion_complete}.
It can be seen that the rendered image from our method was more clear than others.
Furthermore, the depth from our method was smoother and more accurate than others.
Detailed comparisons of different components contributing to the overall performance were discussed in the next section, Ablation Study \ref{sec:abliation_study}.  
}


\begin{table}[t]
	\setlength\tabcolsep{4.5pt}
	\centering
	\caption{{Ablation Study: metric comparison for different components}
		\label{tab:ablation_experiment}}
	\setlength{\tabcolsep}{1mm}
	\begin{tabular}{lccccc}
		\toprule
		\quad & PSNR $\uparrow$ & SSIM $\uparrow$ & LPIPS $\downarrow$ & \begin{tabular}[c]{@{}c@{}}RMSE  $\downarrow$\\{[}m{]} $@1\sigma$ \end{tabular} & \begin{tabular}[c]{@{}c@{}}RMSE  $\downarrow$\\{[}m{]} $@2\sigma$ \end{tabular} \\
		\midrule
		baseline(Nerfacto\cite{nerfstudio}) & 21.33 & 0.748 & 0.193 & 14.13 & 23.84 \\
		baseline+S & \underline{22.47} & \underline{0.753} & \underline{0.186} & 13.56 & 22.95  \\
		baseline+D & 21.54  & 0.750 & 0.191 & \underline{6.89} &  12.76  \\
		baseline+D+O & 21.42 & 0.748 & 0.193 & 7.34 & \underline{12.56} \\
		baseline+S+D+O(Ours) & \textbf{22.54} & \textbf{0.754} & \textbf{0.185} &  \textbf{6.11} & \textbf{11.34} \\
		\bottomrule
	\end{tabular}
	\begin{tablenotes}
		\footnotesize
		\item[1]  *Comparisons of different components.
		S refers to sequence appearance embedding.
		D refers to ground surface supervision.
		O refers to occlusion completion. 
		
	\end{tablenotes}
\end{table}

\subsection{Ablation Study}
\label{sec:abliation_study}
Besides the engineering contribution of the crowd-sourcing framework, three theoretical novel components were proposed: sequence appearance embedding, ground surface supervision, and occlusion completion.
We performed an ablation study to investigate the individual contribution of these components and evaluated their combined effects.
Nerfacto\cite{nerfstudio} was the baseline method, which different components were added on.
The detailed metrics were shown in Table. \ref{tab:ablation_experiment}.
It can be seen that the appearance performance (PSNR, SSIM, and LPIPS) improved by adding sequence appearance embedding.
The depth metric (RMSE) improved a lot by adding ground surface supervision.
The depth error value still seemed large, since only the nearby ground surface was optimized, there was still a big depth error in the distant scene.  
Occlusion completion was not evaluated separately, as it was based on ground surface supervision.
It seemed that occlusion completion had no effect on the quantitative metrics, however, it contributed a lot to the rendered image, as shown in the following qualitative analysis.
In the following, we gave specific examples to illustrate the effect of these components qualitatively:

\subsubsection{\textbf{Sequence Appearance Embedding}}

As shown in Fig. \ref{fig:render_quality}, lacking sequence appearance embedding, the appearance of rendered images from Mip-NeRF and Instant-NGP was far away from the ground truth.
However, the rendered image from our method was similar to ground truth in terms of weather and daytime.


\subsubsection{\textbf{Ground Surface Supervision}}

\begin{figure}[t]
	\centering
	\subfigure[Mesh (Nerfacto).]{
		\includegraphics[width=.45\linewidth]{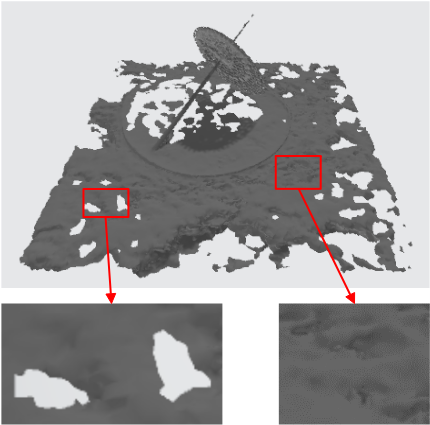} \label{fig:surface_a}
	}
	\subfigure[Mesh (Ours).]{
		\includegraphics[width=.45\linewidth]{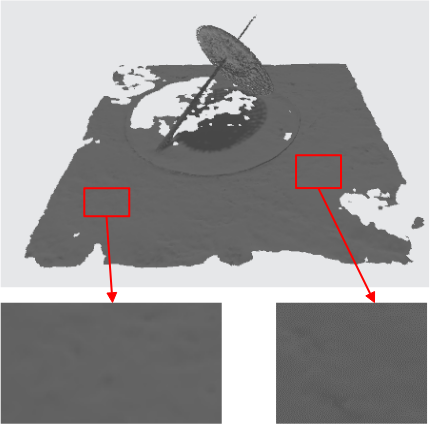} \label{fig:surface_b}
	}
	\subfigure[Point Cloud (Nerfacto).]{
		\includegraphics[width=.45\linewidth]{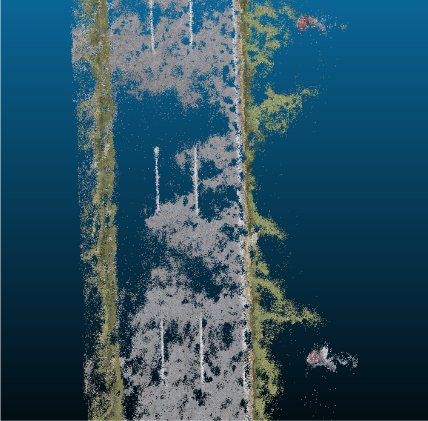} \label{fig:surface_c}
	}
	\subfigure[Point Cloud (Ours).]{
		\includegraphics[width=.45\linewidth]{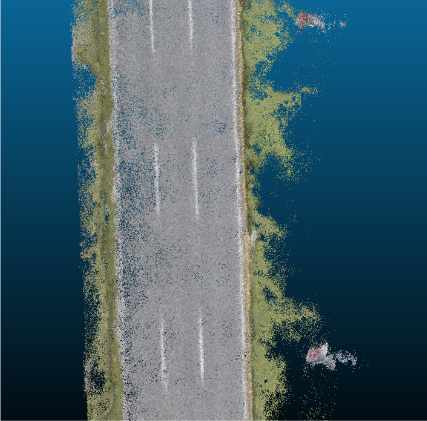} \label{fig:surface_d}
	}
	\caption{Comparisons on surface reconstruction. The road surface of our mehtod was more complete and flatter due to depth supervision and occlusion completion. }
	\label{fig:surface}
	\end {figure}

To visualize depth reconstruction quality intuitively, we explicitly extracted the mesh and point cloud from the implicit NeRF model, as shown in Fig. \ref{fig:surface}. 
We used two formats to represent geometry, which were mesh and point cloud.
\revise{By querying the density of every 3D space point brute-forcely and setting an occupied threshold, the dense point cloud was generated.
Then, Poisson surface reconstruction \cite{kazhdan2006poisson} outputs high-quality meshes.}
It can be seen that there were a lot of holes from Nerfacto, and the surface was irregular, in Fig. \ref{fig:surface_a} and Fig. \ref{fig:surface_c}.
It was the common phenomenon for the NeRF-based method, which achieved poor geometric structure.
However, the ground surface was completer and flatter with the ground depth supervision, as shown in Fig. \ref{fig:surface_b} and Fig. \ref{fig:surface_d}.

\subsubsection{\textbf{Occlusion Completion}}
\begin{figure}[t]
		\centering
		\includegraphics[width=0.45\textwidth]{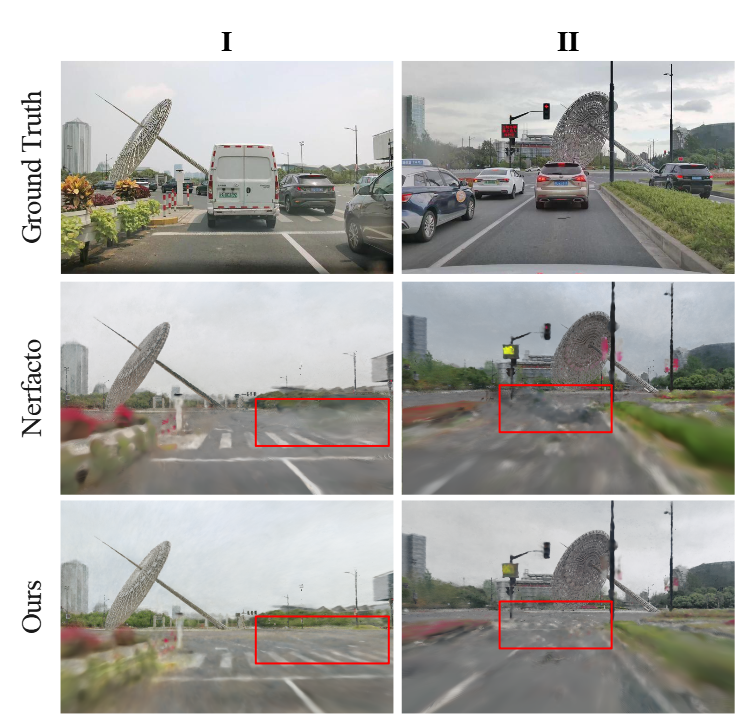}
		\caption{Comparison on render quality near traffic flow. Due to the occlusion completion, the shaded area was predict well in our system.}
		\label{fig:occlusion_complete}
	\end{figure}

Although the metric of rendering quality seemed no improvement in Table \ref{tab:ablation_experiment}, the occluded area (masked by moving objects) was effectively restored in the image view.
Examples were shown in Fig. \ref{fig:occlusion_complete}.
The mask of moving objects resulted in the black hold on the synthetic view in Nerfacto, without occlusion completion.
However, the shaded area on the ground was effectively recovered by our occlusion completion.
Since the masked area was interpolated from its nearby road surface in the training process, our neural network was able to predict the blocked area.
Unfortunately, due to the lack of ground truth of the masked region, the metric number was not improved.

Overall, the combination of these three components, sequence appearance embedding, ground surface supervision, and occlusion completion, contributed to the overall performance.

%% file: sections/experiment_trips.tex

\subsection{Numbers of Trips Comparison}
Since our system is a crowd-sourced framework, the result is affected by the amount of data directly.
With the continuous increase of the crowd-sourced data, the model becomes more and more complete and accurate.
Objectively speaking, the proposed system had great growth potential.
We conducted experiments with different numbers of trips and training images to illustrate this point.
Increasing trips in one scene can improve the coverage, and reduce the ambiguities caused by unseen views.
As shown in Table \ref{tab:experiment_trips}, the proposed system, CS-NeRF, obtained higher performance with more training images from different trips.
We also showed qualitative results of this experiments in Fig. \ref{fig:experiment_trip_vis}.
Fewer floaters in rendered images and more precise rendered depth could be obtained, with the increase of trips.

\begin{table}[t]
    \caption{{Metric comparison for different numbers of trips}
    \label{tab:experiment_trips}}
    \setlength{\tabcolsep}{1mm}
    \centering
    \begin{tabular}{@{}ccccccc@{}}
        \toprule
        trips & \begin{tabular}[c]{@{}c@{}} training\\ images\end{tabular} & PSNR $\uparrow$ & SSIM $\uparrow$ & LPIPS $\downarrow$ & \begin{tabular}[c]{@{}c@{}}RMSE{[}m{]}\\ $@1\sigma$ $\downarrow$\end{tabular} & \begin{tabular}[c]{@{}c@{}}RMSE{[}m{]}\\ $@2\sigma$ $\downarrow$\end{tabular} \\ \midrule
        5           & 539                                                             & 16.73  & 0.676  & 0.272   & 7.33                                                               & 11.90                                                              \\
        10          & 1104                                                            & 16.89  & 0.686  & 0.262   & 8.18                                                               & 17.38                                                              \\
        15          & 1582                                                            & 17.10  & 0.701  & 0.249   & 6.74                                                               & 12.63                                                              \\
        30          & 2104                                                            & 17.24  & \underline{0.705}  & 0.244   & \underline{6.60}                                                               & \underline{11.86}                                                              \\
        45          & 2474                                                            & \underline{17.58}  & \textbf{0.718}  & \underline{0.229}   & 6.66                                                               & 11.92                                                              \\
        58          & 2845                                                            & \textbf{17.79}  & \textbf{0.718}  & \textbf{0.226}   & \textbf{6.13}                                                               & \textbf{11.54}                                                              \\ \bottomrule
    \end{tabular}
\end{table}


%% file: sections/experiment_navigation.tex

\begin{figure*}
	\centering
	\includegraphics[width=0.9\textwidth]{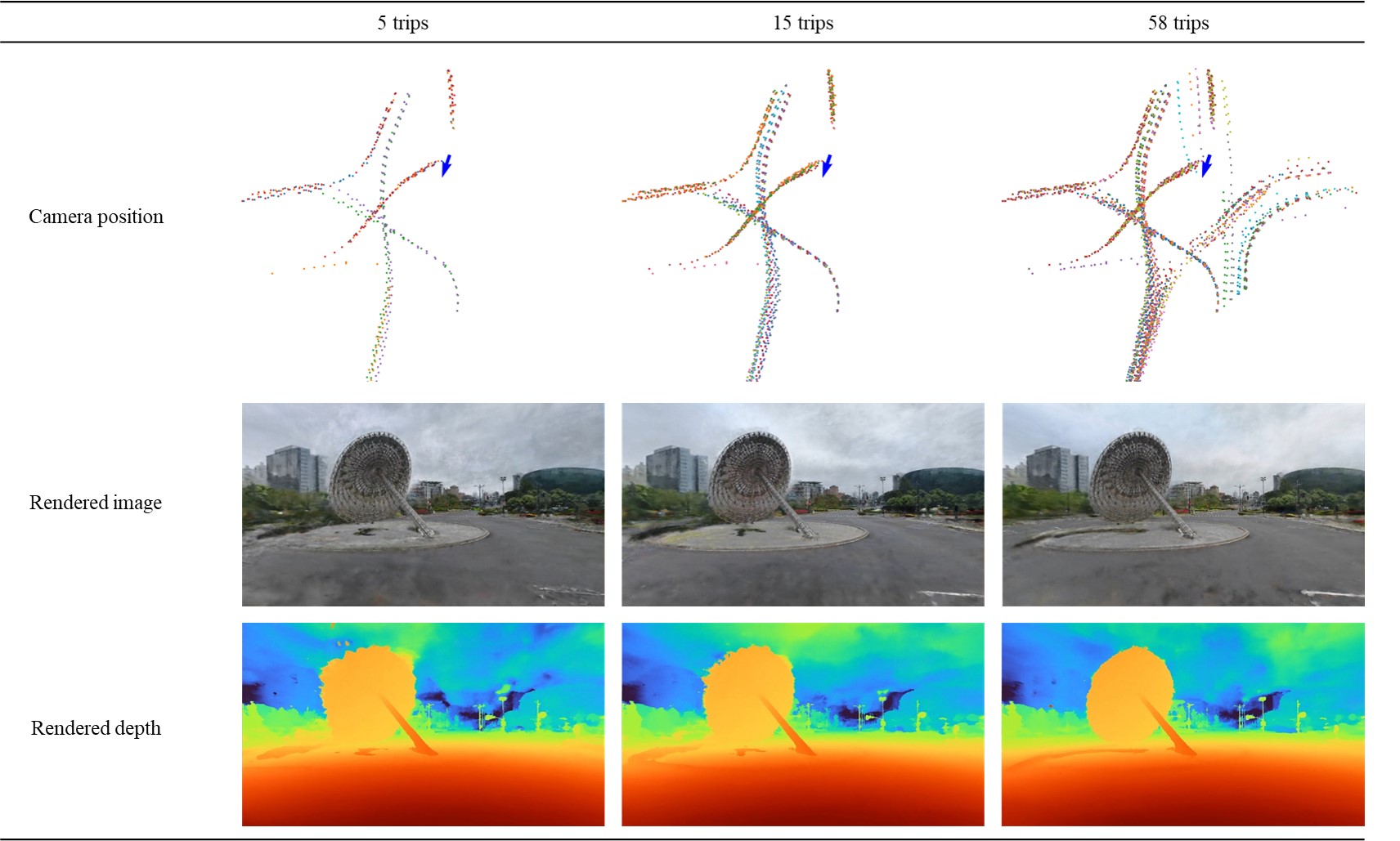}
	\caption{Qualitative comparison on different trips.
			In the first row, the colorful dots stands for the view point in the X-Y plane.
			The blue arrow indicates the view of rendering.
			The rendered images and depths are shown in the second and third rows, respectively.
		It can be seen that the scene reconstruction becomes more complete and precise with more trips.  }
	\label{fig:experiment_trip_vis}
\end{figure*}


%% file: sections/method_rendering.tex
\section{Application: 3D Navigation}
\label{sec:rendering}
\revv{
By leveraging the NeRF model, we could render novel views with any virtual elements.
In this section, we rendered the image with a guideline in this complex traffic intersection, which could provide users with an first-view navigation experience.
As shown in Fig. \ref{fig:abs_b}, a reference line is rendered on the view to guide the driver vividly.
We elaborate on how to render a novel view with guidance markers to achieve first-view navigation.
}

\begin{figure}[t]
	\centering
	\includegraphics[width=0.4\textwidth]{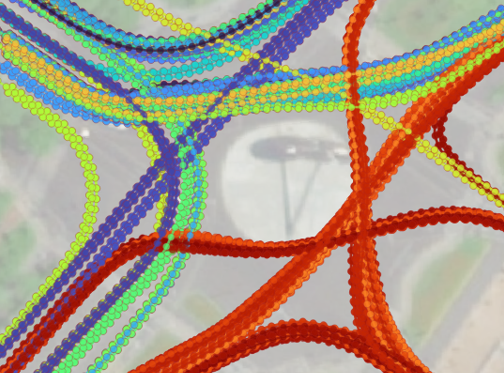}
	\caption{The illustration of crowd-souring human's driving paths within one intersection. These driving paths can be used to guide followers.}
	\label{fig:trajectory}
\end{figure}

\revv{
\subsection{Guide Marker Generation}
In addition to images, trajectories of vehicles were collected through the crowd-sourcing platform.
As shown in Fig. \ref{fig:trajectory}, multiple human's driving paths were collected within one intersection.
The driving path from experienced drivers can be used to guide new-hand drivers.
Therefore, we picked up some smooth trajectories, and treated them as guiding markers of the route in the complicated crossroad.
}



\begin{figure}[t]
	\label{alg:NIG}
	\renewcommand{\algorithmicrequire}{\textbf{Input:}}
	\renewcommand{\algorithmicensure}{\textbf{Output:}}
	\begin{algorithm}[H]
		\caption{Novel Image Rendering with Guidance Markers}
		\label{rendering_algorithm_pseudo_code}
		\begin{algorithmic}[1]
			\REQUIRE Trained model $\mathbb{M}$, Guidance Trajectory $\mathbb{P}$ 
			\ENSURE Navigation Image $\mathbb{I}$  
			
			\STATE Generate the polygon area of guidance markers $E=\{E_1, E_2, ..., E_k\}$ from trajectory $\mathbb{P}$;
			
			\FOR{each ray $r$ in $\mathbb{I}$}
			\STATE Color $c_r$, depth $d_r$ $\leftarrow$ Query $r$ in $\mathbb{M}$;
			\FOR{$j=1$ to $k$}
				\IF{$r$ intersects with $E_j$}
				\STATE $D \leftarrow$ Add the depth of intersection point;
				\ENDIF 
			\ENDFOR
			\IF{$min(D) < d_r$}
			\STATE $c \leftarrow$ Compose scene color $c_r$ with marker color $c_m$;
			\ELSE
			\STATE $c \leftarrow$ Scene color $c_r$; 
			\ENDIF 
			\STATE Update color $c$ to navigation image $\mathbb{I}$;
			\ENDFOR
			
		\end{algorithmic}
	\end{algorithm}
\end{figure}



\begin{figure}[t]
	\centering
	\subfigure[The illustration of rendering novel view with navigation information.]{
		\includegraphics[width=.9\linewidth]{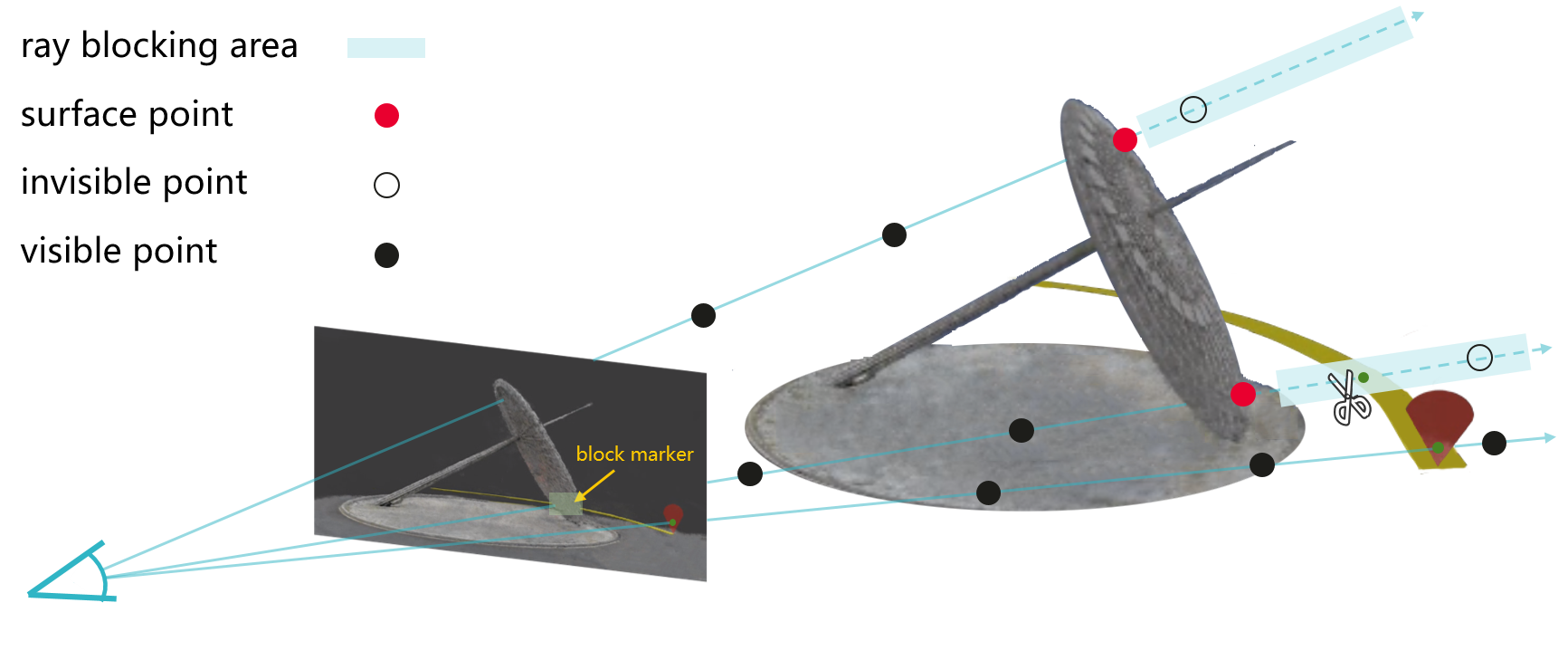} \label{fig:rendering_navigation_view}
	}
	\subfigure[The Example of rendered view with occlusion.]{
		\includegraphics[width=.8\linewidth]{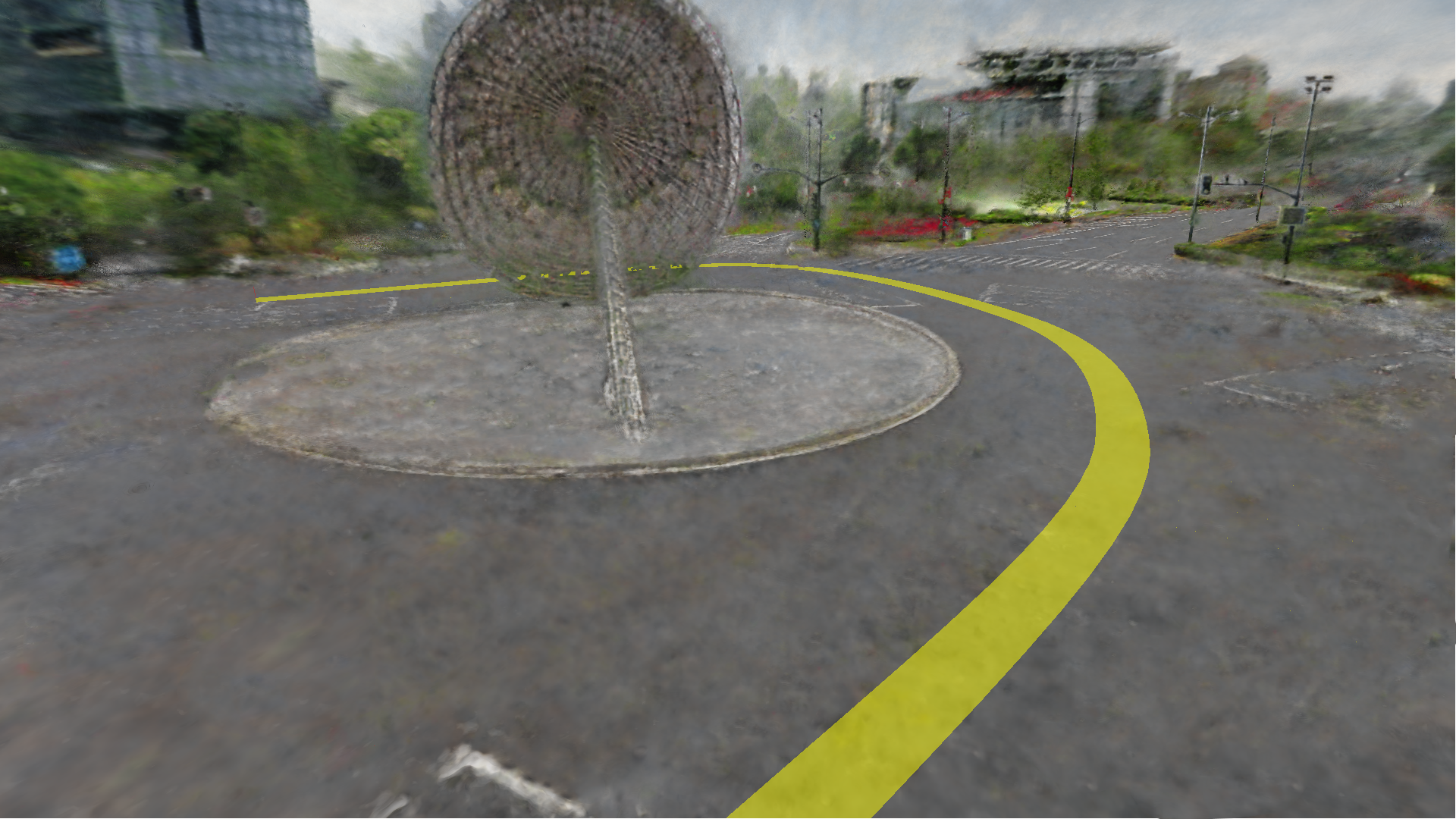} 
		\label{fig:rendering_occlusion}
	}
	\caption{(a) The illustration of rendering novel view with navigation information. (b) The Example of rendered view with occlusion.}
	\label{fig:rendering}
\end{figure}

\revv{
\subsection{Novel View Rendering}
In the following, we elaborate on how to render the novel view image with guidance markers.
The whole procedures are detailed in pseudo-code in Algorithm \ref{rendering_algorithm_pseudo_code}.
Firstly, as same with the normal rendering process, eq.(\ref{eq:color_render}) is performed on every ray. 
Secondly, we add the marker's color on this pixel if the ray intersects with the marker.
The overlaid color is,
\begin{equation}
	C^{'}(\mathbf{r}) = \alpha C(\mathbf{r}) + (1-\alpha) \mathbf{c}
\end{equation}
where $\mathbf{c}$ is the constant color value of the maker, $\alpha$ is a constant scale for alpha composition. 
We use $\alpha = 0.3$ in the implementation to achieve a transparent appearance.
}

\revv{
However, in some places, the guidance marker is occluded by the scene.
Therefore, we need to deal with occlusion specially.
As shown in Fig. \ref{fig:rendering_navigation_view}, if the ray is terminated before the maker, in other words, the depth of the ray is shorter than the marker, we determine occlusion happens.
The ray will keep the original color.
\revv{Otherwise, the ray will be overlaid with the marker’s color.}
An example of a novel view with an occluded marker was shown in Fig. \ref{fig:rendering_occlusion}. 
In this way, we can achieve an first-view navigation experience, meanwhile, the driver can change view angles online freely. 
}

\revv{
As shown in Fig. \ref{fig:immersive_navigation_compare_image}, we compared it with the traditional navigation tool, Amap.
Amap only provided users with the 2D picture at a fixed angle view.
However, our system could provide users with a realistic picture, which was similar to the real scene.
In addition, the viewing angle could be changed arbitrarily, such as the front view, and top-down view, to give the user an clearer experience.
}

%% file: sections/conclusion.tex
\section{Conclusion}
\label{sec:conclusion}
In this paper, we proposed a crowd-sourced framework that trained the NeRF model from the data captured by multiple production vehicles.
This approach solved a key problem of large-scale reconstruction, that was where the data came from.
We incorporated multiple improvements, such as ground surface supervision, occlusion completion, and sequence appearance embedding, to enhance the performance.
Finally, the 3D first-view navigation based on the NeRF model was applied to real-world scenarios.

Although the result from the proposed CS-NeRF framework seems great and promising, there are still several limitations and future works that are worth discussing:

\begin{enumerate}
\item Realistic Changes: 
Handling realistic changes (temporal inconsistency) is challenging during NeRF reconstruction. 
Temporary changes (e.g., lane redrawing) and long-term changes (e.g., road expansions) need to be accurately identified and incorporated into the reconstruction process.
\item Privacy Concerns: 
Due to the nature of user data, privacy concerns need to be addressed. 
Appropriate data processing and anonymization measures are implemented to ensure user privacy and data security.
\item Data Quality and Sensor Variations: 
Variations in data quality and sensor types across different platforms impact the accuracy of NeRF reconstruction. 
Data preprocessing, standardization, and calibration techniques are employed to mitigate these variations and ensure consistent data quality.
\end{enumerate}

In summary, the paper proved the concept that crowd-sourcing way can be applied to the real world.
In the future, the work will be extended to the automotive industry, where it can obtain a mass of driving data.
Meanwhile, the NeRF model will contribute to the development of autonomous driving.

 \begin{figure}[t]
	\centering
	\includegraphics[width=0.45\textwidth]{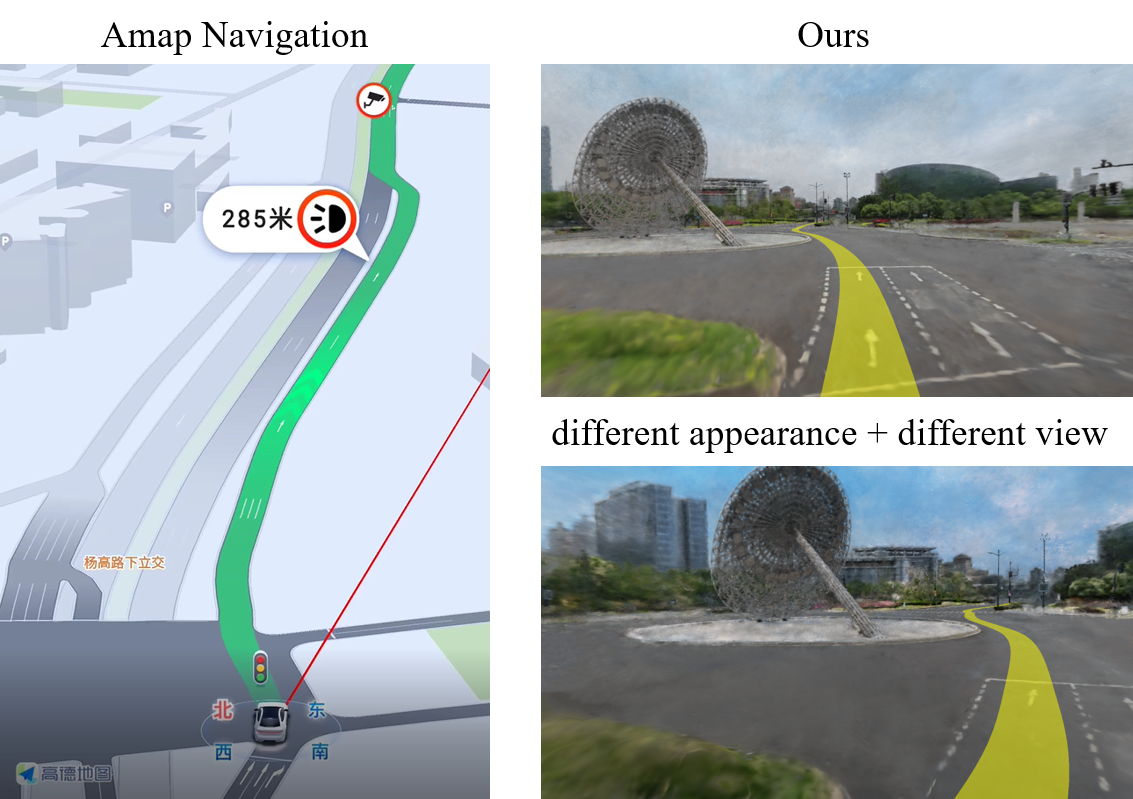}
	\caption{ The navigation application of the proposed system compared against Amap. Traditional navigation tools only provide 2D navigation pictures, while our approach can provide 3D images with different appearance styles and with different angles of view.}
	\label{fig:immersive_navigation_compare_image}
\end{figure} 

%% file: sections/biography.tex
\begin{IEEEbiography}[{\includegraphics[width=1in,height=1.25in,clip,keepaspectratio]{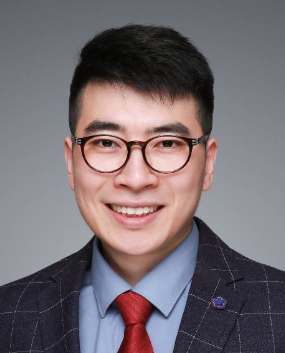}}]{Tong Qin}
	received the B.Eng degree in control science and engineering from the Zhejiang University, China, in 2015, and the Ph.D. degree in the Department of Electronic and Computer Engineering, the Hong Kong University of Science and Technology, HongKong, in 2019.
	He worked as a staff research scientist in the department of Advanced Driving Solution, Huawei from 2019 to 2023. 
	He is currently working as an associate professor in Global Institute of Future Technology, Shanghai Jiao Tong University.
	His research interests include SLAM, NeRF, machine learning, and autonomous driving.
\end{IEEEbiography}

\begin{IEEEbiography}[{\includegraphics[width=1in,height=1.25in,clip,keepaspectratio]{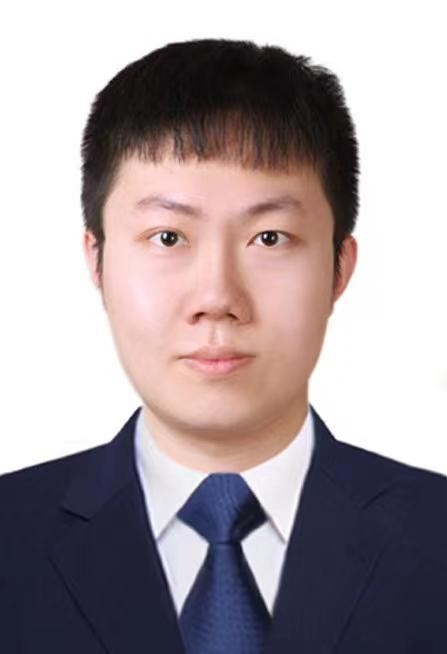}}]{Changze Li}
	received the B.Eng. degree in control science and engineering from Xidian University, China, in 2019, and the M.S. degree in Navigation Guidance and Control, the Northwestern Polytechnical University, China, in 2022.
	He worked as a research engineer in the department of Advanced Driving Solution, Huawei. 
	He is currently a PhD candidate in Shanghai Jiao Tong University.
	His research interests include NeRF, computer vision in the field of autonomous driving.
\end{IEEEbiography}

\begin{IEEEbiography}[{\includegraphics[width=1in,height=1.25in,clip,keepaspectratio]{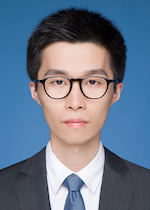}}]{Haoyang Ye}
	received the B.Eng. degree in automation from the College of Control Science and Engineering, Zhejiang University, China, in 2016, and the Ph.D. degree from the Department of Electronic and Computer Engineering, Hong Kong University of Science and Technology, Hong Kong, in 2020.
	
	He is currently working as a Research Engineer with Shanghai Huawei Technologies Co., Ltd., Shanghai, China.
	His research interests include state estimation for robotics, SLAM, sensor fusion, and computer vision.
\end{IEEEbiography}

\begin{IEEEbiography}[{\includegraphics[width=1in,height=1.25in,clip,keepaspectratio]{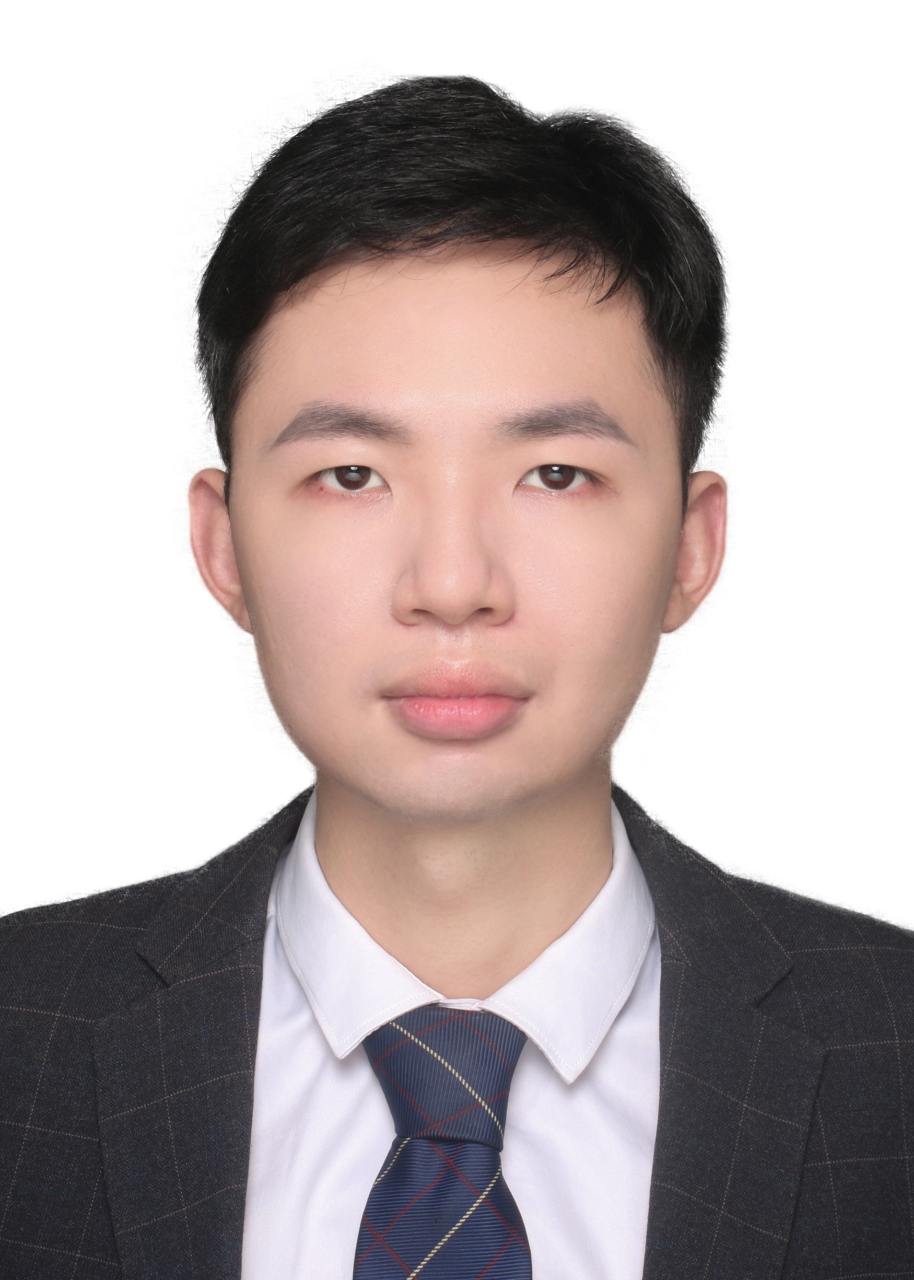}}]{Shaowei Wan}
	received the B.Eng. degree in Robot Engineering from Northeastern University, China, in 2020, and the M.S. degree in Control Engineering, Huazhong University of Science and Technology, China, in 2022.
	
	He is currently working as a research engineer in the department of Advanced Driving Solution, Huawei.  His research interests include NeRF and autonomous driving.
\end{IEEEbiography}

\begin{IEEEbiography}[{\includegraphics[width=1in,height=1.25in,clip,keepaspectratio]{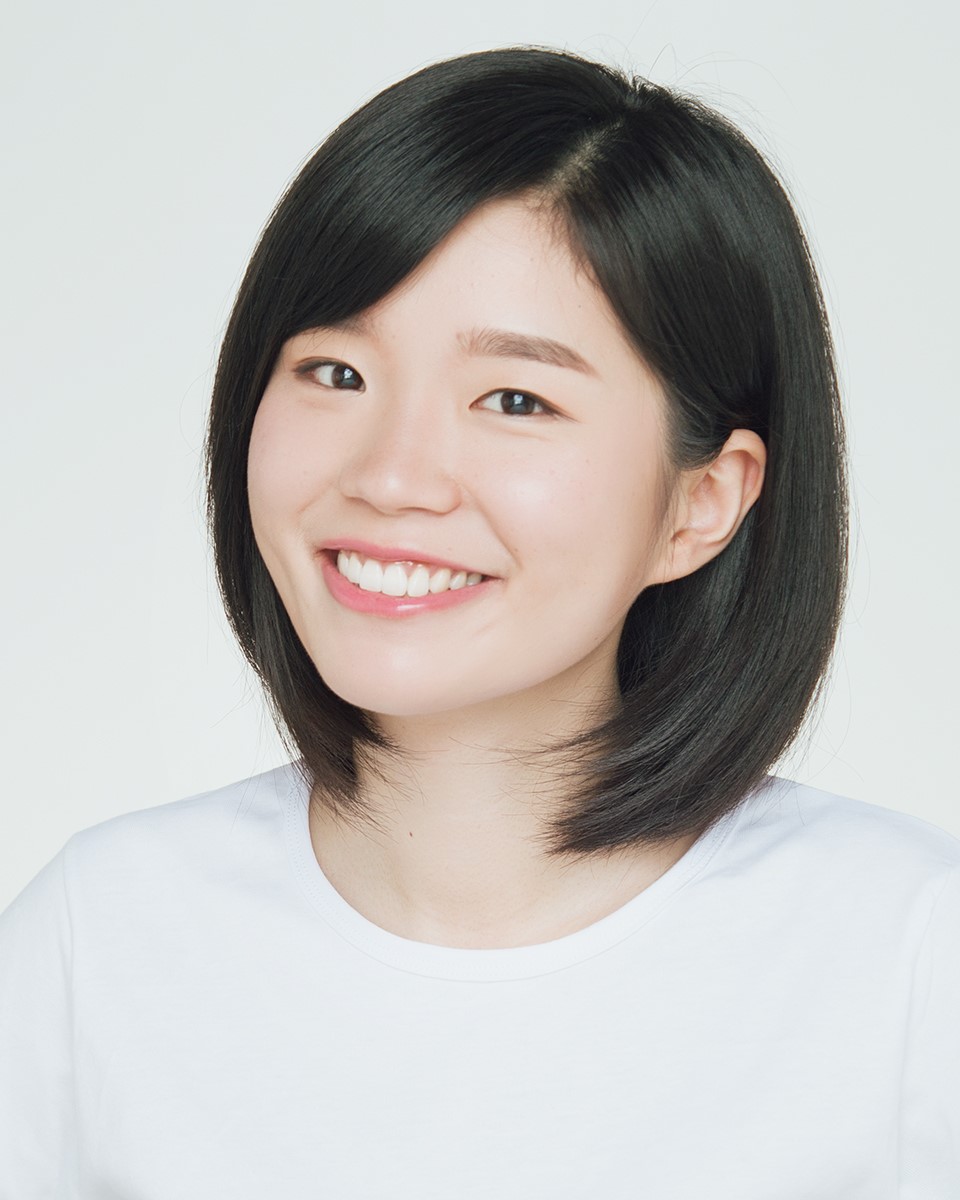}}]{Minzhen Li}
	received her B.Eng. degree in Geographic Information Systems from Tongji University in 2013, followed by her M.S. degree in Surveying Engineering from the same university in 2016.
	
	She is a research engineer working at Huawei in the field of autonomous driving since 2016. Her research interests include Mapping \& Localization, Auto labelling in the field of autonomous driving.
\end{IEEEbiography}


\begin{IEEEbiography}[{\includegraphics[width=1in,height=1.25in,clip,keepaspectratio]{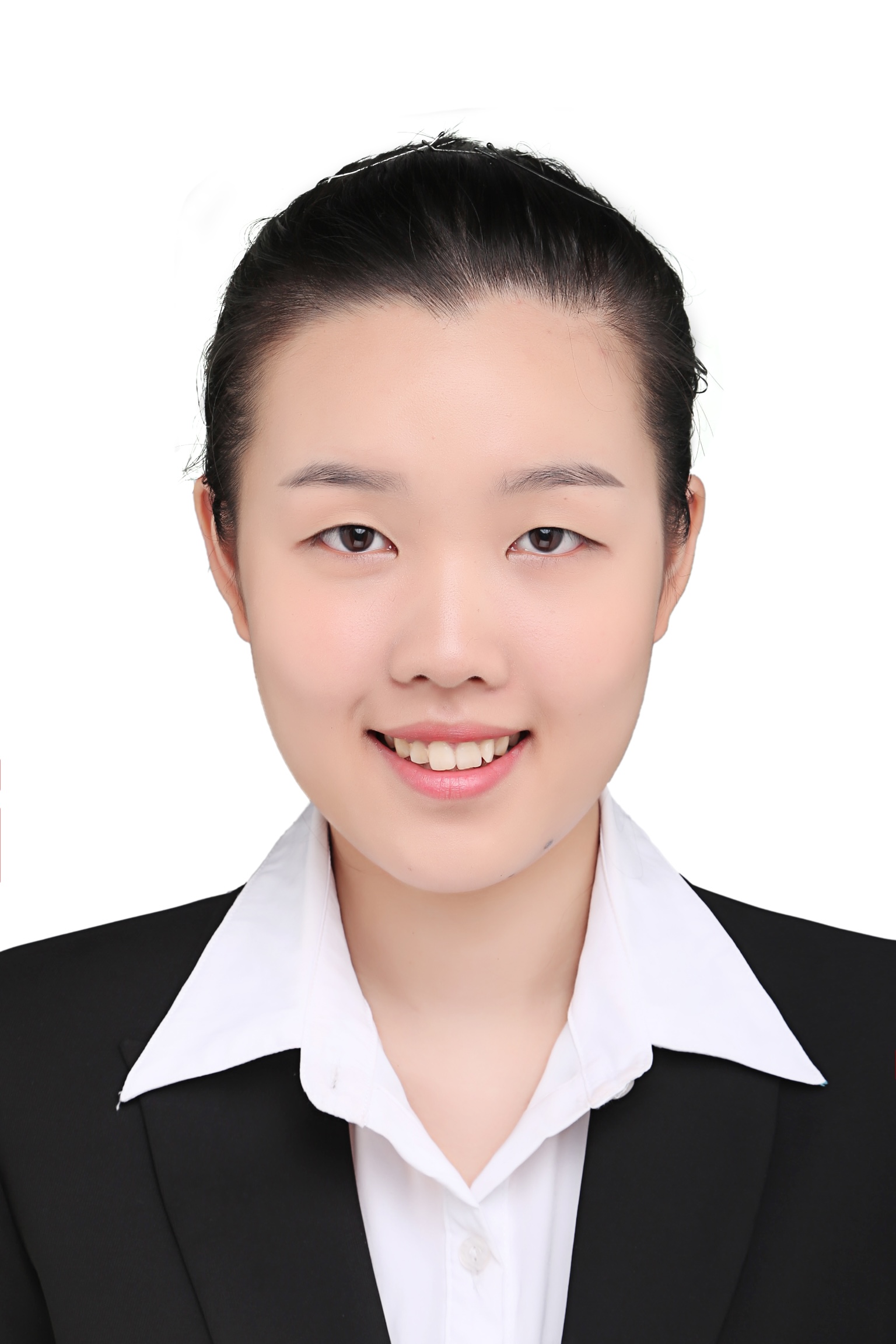}}]{Hongwei Liu}
	received the B.Eng degree in Automobile Engineering from Chongqing Jiaotong University, China, in 2017, and the M.S. degree in Automobile Engineering, Tongji University, China, in 2020.
	
	She is currently working as a research engineer in the department of Advanced Driving Solution, Huawei.  Her research interests include SLAM and autonomous driving.
\end{IEEEbiography}

\begin{IEEEbiography}[{\includegraphics[width=1in,height=1.25in,clip,keepaspectratio]{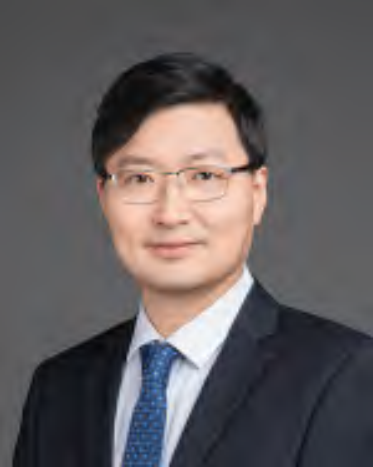}}]{Ming Yang}
received his Master’s and Ph.D. degrees from Tsinghua University, Beijing, China, in1999 and 2003, respectively. Presently, he holdsthe position of Distinguished Professor at ShanghaiJiao Tong University, also serving as the Directorof the Innovation Center of Intelligent ConnectedVehicles. Dr. Yang has been engaged in the researchof intelligent vehicles for more than 25 years.
\end{IEEEbiography}